\documentclass{article}

\usepackage[preprint]{neurips_2023}

\usepackage[utf8]{inputenc} 
\usepackage[T1]{fontenc}    
\usepackage{hyperref}       
\usepackage{url}            
\usepackage{booktabs}       
\usepackage{amsfonts}       
\usepackage{nicefrac}       
\usepackage{microtype}      
\usepackage{xcolor}         

\usepackage{subcaption}

\usepackage{multirow}
\usepackage{amsmath}
\usepackage{capt-of}
\usepackage{tabularx}
\usepackage{epsfig}
\usepackage{amssymb}
\usepackage{amsfonts}
\usepackage{booktabs}
\usepackage{scalerel}
\usepackage[inline]{enumitem}
\usepackage{listings}
\usepackage{varwidth}
\usepackage[export]{adjustbox}
\usepackage{tikz}
\usetikzlibrary{tikzmark}
\usepackage{cleveref}

\usepackage{stmaryrd}
\usepackage{bbm}

\usepackage{algorithm}
\usepackage[noend]{algpseudocode}

\definecolor{deepblue}{rgb}{0,0,0.5}
\definecolor{officeblue}{RGB}{0,102,204}
\definecolor{deepred}{rgb}{0.6,0,0}
\definecolor{deepgreen}{rgb}{0,0.5,0}
\definecolor{mybrickred}{RGB}{182,50,28}

\definecolor{fillcolor}{RGB}{216,217,252}

\algnewcommand\algorithmicrequireb{{\hspace{0.85cm}}}
\algnewcommand\INPTDESCB{\item[\algorithmicrequireb]}

\algnewcommand\algorithmicfuncdesc{\textbf{Function:}}
\algnewcommand\FUNCDESC{\item[\algorithmicfuncdesc]}
\algnewcommand\algorithmicfuncdescb{{\hspace{1.48cm}}}
\algnewcommand\FUNCDESCB{\item[\algorithmicfuncdescb]}
\algnewcommand{\algorithmicgoto}{\textbf{goto}}
\algnewcommand{\Goto}[1]{\algorithmicgoto~\ref{#1}}

\usepackage{amsmath,amsfonts,bm}

\def\eqref#1{equation~\ref{#1}}

\def\1{\bm{1}}

\def\vtheta{{\bm{\theta}}}

\DeclareMathAlphabet{\mathsfit}{\encodingdefault}{\sfdefault}{m}{sl}
\SetMathAlphabet{\mathsfit}{bold}{\encodingdefault}{\sfdefault}{bx}{n}

\newcommand{\Ls}{\mathcal{L}}

\newcommand\our{\textsc{IGap}}
\newcommand\strain{\mathcal{S}_\text{train}}
\newcommand\sval{\mathcal{S}_\text{val}}

\newcommand\tval{\mathcal{T}_\text{val}}

\usepackage{pifont}
\title{Measuring Cross-Lingual Transferability of Multilingual Transformers on Sentence Classification}

\author{Zewen Chi,~~Heyan Huang\thanks{\ \ Corresponding author.},~~Xian-Ling Mao\\
School of Computer Science and Technology, Beijing Institute of Technology\\
\texttt{\{czw,hhy63,maoxl\}@bit.edu.cn}\\}

\begin{document}
\maketitle
\begin{abstract}
Recent studies have exhibited remarkable capabilities of pre-trained multilingual Transformers, especially cross-lingual transferability. However, current methods do not measure cross-lingual transferability well, hindering the understanding of multilingual Transformers. In this paper, we propose \our{}, a cross-lingual transferability metric for multilingual Transformers on sentence classification tasks. \our{} takes training error into consideration, and can also estimate transferability without end-task data. Experimental results show that \our{} outperforms baseline metrics for transferability measuring and transfer direction ranking. Besides, we conduct extensive systematic experiments where we compare transferability among various multilingual Transformers, fine-tuning algorithms, and transfer directions. More importantly, our results reveal three findings about cross-lingual transfer, which helps us to better understand multilingual Transformers. 

\end{abstract}

\section{Introduction}
\label{sec:intro}

The development of Transformer~\citep{transformer} greatly advances natural language processing. In particular, multilingual Transformers such as XLM-R~\citep{xlmr} have become the foundational element of a wide range of natural language processing systems, because of the remarkable cross-lingual abilities they have.

The most appealing ability of multilingual Transformers is cross-lingual transferability. As an early implementation of multilingual Transformer, Multilingual BERT (mBERT;~\citealt{bert}) is pre-trained on Wikipedia text of 104 languages with a single Transformer encoder.
\citet{wu2019beto} fine-tune mBERT on downstream NLP tasks in a single source language, and find that the fine-tuned mBERT can directly perform the tasks in target languages, i.e., zero-shot cross-lingual transfer. Cross-lingual transferability is observed from not only mBERT, but also other multilingual Transformers \citep{xlm,xlmr,infoxlm}. To improve cross-lingual transfer performance, follow-up studies have tried to develop various pre-trained multilingual Transformers \citep{infoxlm,hictl,mt5}, or activate the transferability by designing fine-tuning algorithms \citep{filter,xtune}. Despite the success, current methods do not measure cross-lingual transferability well,
leading to difficulty in analyzing and understanding multilingual Transformers.

In this paper, we explore cross-lingual transferability metrics of pre-trained multilingual Transformers on sentence classification tasks. We propose \our{}, a cross-lingual transferability metric that allows us to compare the transferability among multilingual Transformers, fine-tuning algorithms, and transfer directions. 
Specifically, we decompose the cross-lingual transfer error into three parts, which are interlingual transfer gap, intralingual generalization gap, and training error. \our{} measures transferability by searching the minimum interlingual transfer gap while also taking training error into consideration. Moreover, by transferring randomly-generated labels, \our{} can estimate cross-lingual transferability when end-task data are unavailable.

To validate \our{}, we compare \our{} with baseline metrics for transferability measuring. We also present a new task to evaluate transferability metrics, called transfer direction ranking, where the goal is to predict which target language obtains better transfer performance for a specific source language. The evaluation results demonstrate that \our{} better indicates cross-lingual transferability of multilingual Transformers than both the cross-lingual transfer gap metric \citep{xtreme} and representation-based metrics. Besides, our experimental results demonstrate that \our{} can estimate cross-lingual transferability when downstream task data are unavailable. Furthermore, to better understand multilingual Transformers, we conduct extensive systematic experiments, where we compare cross-lingual transferability among multilingual Transformers, fine-tuning algorithms, and transfer directions. Through empirical analysis, we have three findings: (1) Cross-lingual transfer methods implicitly reduce \our{}. (2) Multilingual Transformers can memorize new knowledge and transfer it to other languages. (3) Better-aligned representations do not promise better cross-lingual transferability.

Our contributions are as follows:
\begin{itemize}
\item We propose \our{}, a cross-lingual transferability metric for multilingual Transformers on sentence classification tasks.
\item We conduct systematic comparisons of cross-lingual transferability among multilingual Transformers, fine-tuning algorithms, and transfer directions.
\item We present the transfer ranking direction task to evaluate transferability metrics.
\item We reveal three findings about cross-lingual transfer, which help us to better understand multilingual Transformers.
\end{itemize}

\section{Background}
\label{sec:back}

\subsection{Cross-lingual transfer}

We focus on the zero-shot  cross-lingual transfer of BERT-style multilingual Transformers~\citep{bert,xlm,xlmr}. 
Let $\vtheta_0$ denote the pre-trained multilingual Transformer that will be fine-tuned on a downstream sentence classification task, which we refer to as \textit{end task} for simplicity.
The goal of cross-lingual transfer is to transfer the end-task knowledge from a source language to target languages, which is commonly achieved by fine-tuning $\vtheta_0$ on the end task in the source language $s$:
\begin{align}
    \arg\min_{\vtheta} \sum_{(x, y) \sim \strain} \Ls (x, y; \vtheta)
\end{align}
where $\strain$ and $\Ls$ are the training set and the loss function of the end task, and the model $\vtheta$ is initialized from the pre-trained model $\vtheta_0$. After fine-tuning, the model can directly perform the end task in the target language $t$.

\subsection{Cross-lingual transfer gap}

The end-task result in the target language is a common indicator of cross-lingual transfer performance. However, the end-task results not only indicate cross-lingual transferability but also other capabilities such as representation quality. Therefore, several studies have tried to measure cross-lingual transferability by cross-lingual transfer gap metric~\citep{xtreme,filter,yang2022enhancing}, which is computed by subtracting the performance on target-language validation sets from the performance on the source-language validation set. Let $\sval$ and $\tval$ denote the validation sets in the source and target languages, respectively. Cross-lingual transfer gap can be written as
\begin{align}
\begin{split}
    \mathcal{G}_\text{gap} =&~ \frac{1}{|\sval|} \sum_{(x,y)\sim \sval} \mathcal{M}(x, y; \vtheta) - \frac{1}{|\tval|} \sum_{(x,y)\sim \tval} \mathcal{M}(x, y; \vtheta)
\end{split}
\end{align}
where $\mathcal{M}$ represents the metric that measures end-task performance. For example, the accuracy metric is used as $\mathcal{M}$ for text classification tasks. In what follows, cross-lingual transfer gap is also called transfer gap for simplicity.
In our experiments, we will show how transfer gap fails to measure transferability (Section~\ref{sec:exp1} and Section~\ref{sec:ctd}).

\section{Measuring cross-lingual transferability}

In this section, we first present the design principles that cross-lingual transferability metrics should follow. Then, following the principles, we propose our cross-lingual transferability metric, \our{}.

\subsection{Design principles}

\paragraph{The difficulty of cross-lingual transfer varies with end-task difficulty.} The transfer difficulty varies in two aspects. First, end tasks have various goals, leading to various transfer difficulty, i.e., we should not compare cross-lingual transferability across tasks. Second, even on the same end task, the difficulty varies when we learn the task to varying degrees of proficiency. Therefore, the learning degree should be taken into consideration when developing a cross-lingual transferability metric.

\paragraph{Transferability should be comparable among models, training algorithms, and transfer directions.} Our metric is designed to produce comparable scores, which enable us to compare cross-lingual transferability among models, training algorithms, and transfer directions.

\paragraph{Carefully handle negative transferability.} 
If the model has already learned the end task in some target languages before, we allow the transferability score to be negative, because the end-task knowledge can be transferred from some of the target languages to the source language at the beginning.
Under the zero-shot cross-lingual transfer setting, the end-task knowledge is transferred from the source language to target languages, which leads to positive transferability. In this work, we focus on the zero-shot transfer setting.

\subsection{\our{}}
\label{sec:igap}

Consider a pre-trained multilingual model $\vtheta_{0}$, which is fine-tuned to perform an end task. Under the cross-lingual transfer setting, the model is fine-tuned on $\strain$, which is in a source language $s$, and then evaluated in a target language $t$. 
The empirical cross-lingual transfer error is defined as
\begin{align}
    \mathcal{E} = \frac{1}{|\tval|} \sum_{(x,y)\sim \tval} \mathbbm{1} \{ y \neq \hat{y}_x \},
\end{align}
where $\tval$ stands for the validation set in the target language $t$, and $x, y$ denotes the input text and the golden label, respectively. $\hat{y}_x$ denotes the output label predicted by the end-task model $\vtheta$ fine-tuned from $\vtheta_0$.
The empirical cross-lingual transfer error directly reflects the end-task performance, which depends on not only transferability but also other factors as mentioned above. 
We decompose the cross-lingual transfer error into three components:
\begin{align}
\begin{split}
\mathcal{E} =&~ \mathcal{G}_\text{inter} + \mathcal{G}_\text{intra} + \mathcal{E}_\text{train} \\
\mathcal{G}_\text{inter} =&~ \frac{1}{|\strain|} \sum_{(x,y)\sim \strain} \mathbbm{1} \{ y \neq \hat{y}_{x_t} \} - \mathbbm{1} \{ y \neq \hat{y}_{x} \} \\ 
\mathcal{G}_\text{intra} =&~ \frac{1}{|\tval|} \sum_{(x,y)\sim \tval} \mathbbm{1} \{ y \neq \hat{y}_{x} \} - \frac{1}{|\strain|} \sum_{(x,y)\sim \strain} \mathbbm{1} \{ y \neq \hat{y}_{x_t} \} 
\end{split}
\label{eq:inter}
\end{align}
where $x_t$ stands for the human translation of $x$ from language $s$ into language $t$. 
We name the three terms $\mathcal{G}_\text{inter}$, $\mathcal{G}_\text{intra}$, and $\mathcal{E}_\text{train}$ as interlingual transfer gap, intralingual generalization gap, and training error, respectively.

\textbf{Interlingual transfer gap} $\mathcal{G}_\text{inter}$ measures how much knowledge is lost after the cross-lingual transfer. Notice that we use the translated examples $x_t$ rather than using examples from $\tval$, which ensures the examples in the target language have the same end-task difficulty as the training examples in the source language. Thus, when $\mathcal{G}_\text{inter}$ compares the error between the two languages, it indicates how much end-task knowledge is lost in the target language.
\textbf{Intralingual generalization gap} $\mathcal{G}_\text{intra}$ is the second term of Equation~(\ref{eq:inter}). $\mathcal{G}_\text{intra}$ measures the ability of the model to generalization on unseen examples within the same language. 
\textbf{Training error} $\mathcal{E}_\text{train}$ measures how well the model learns the end task on the training set.
Although interlingual transfer gap reflects cross-lingual transferability, it does not satisfy the design principles because the end-task difficulty is not considered. Thus, \our{} considers the end-task difficulty by comparing transferability under specific training errors.

\paragraph{Measure transferability} Consider a pre-trained multilingual Transformer model $\vtheta_0$. With a specific fine-tuning algorithm $\mathcal{A}$, we can obtain a set of fine-tuned models $\mathcal{A}(\vtheta_0)$ using various hyperparameters including training steps and random seeds. We would like to quantify transferability under a specific training error $\mathcal{E}'$. However, it is intractable to control the model to be fine-tuned to an arbitrary training error precisely. Hence, we use a relaxed condition where we allow the training error of the fine-tuned models to be a little larger than $\mathcal{E}'$, controlled by a fixed error term $\epsilon$. The \our{} metric is computed by:
\begin{align}
    \our{}(\mathcal{E}') = \min_{\genfrac{}{}{0pt}{2}{\vtheta \in \mathcal{A}(\vtheta_0)}{ 0 \leq \mathcal{E}_\text{train} - \mathcal{E}' < \epsilon}} \{ \mathcal{G}_\text{inter} \}.
\label{eq:igap}
\end{align}
$\mathcal{A}(\vtheta_0)$ can be obtained by fine-tuning the pre-trained models with various random seeds and saving intermediate checkpoints frequently. Empirically, \our{} produces positive transferability scores under the zero-shot transfer setting, which is validated in our experiments (Section~\ref{sec:exp1}).

\paragraph{Estimate transferability without end-task data} It is a common situation that end-task data are unavailable for low-resource languages. Different from cross-lingual transfer gap which relies on end-task data, \our{} can also estimate transferability when end-task data are unavailable. We first construct datasets with randomly-generated ``knowledge'' to be transferred. Specifically, we use a parallel corpus $\{(x, x_t)\}$ as the base dataset, and then we generate random $0/1$ labels as the knowledge to be transferred, i.e., $y \sim~ \text{Bernoulli}(1/2)$,
where we ensure that parallel sentences have the same labels.
The second step is to obtain $\mathcal{A}(\vtheta_0)$ by fine-tuning the models on the generated data. Finally, we compute the \our{} scores by Equation (\ref{eq:igap}). Using randomly-generated labels has two advantages. First, it is guaranteed that randomly-generated labels are not seen by the pre-trained models. Second, compared to task-specific data, parallel sentences are typically much easier to obtain for low-resource languages. Empirically, we show that \our{} can effectively select the source languages for cross-lingual transfer (Section~\ref{sec:ctd}).

\begin{figure*}[t!]
\centering
\begin{subfigure}[b]{1.0\textwidth}
    \begin{subfigure}[b]{0.32\textwidth}
     \centering
     \includegraphics[width=0.95\textwidth]{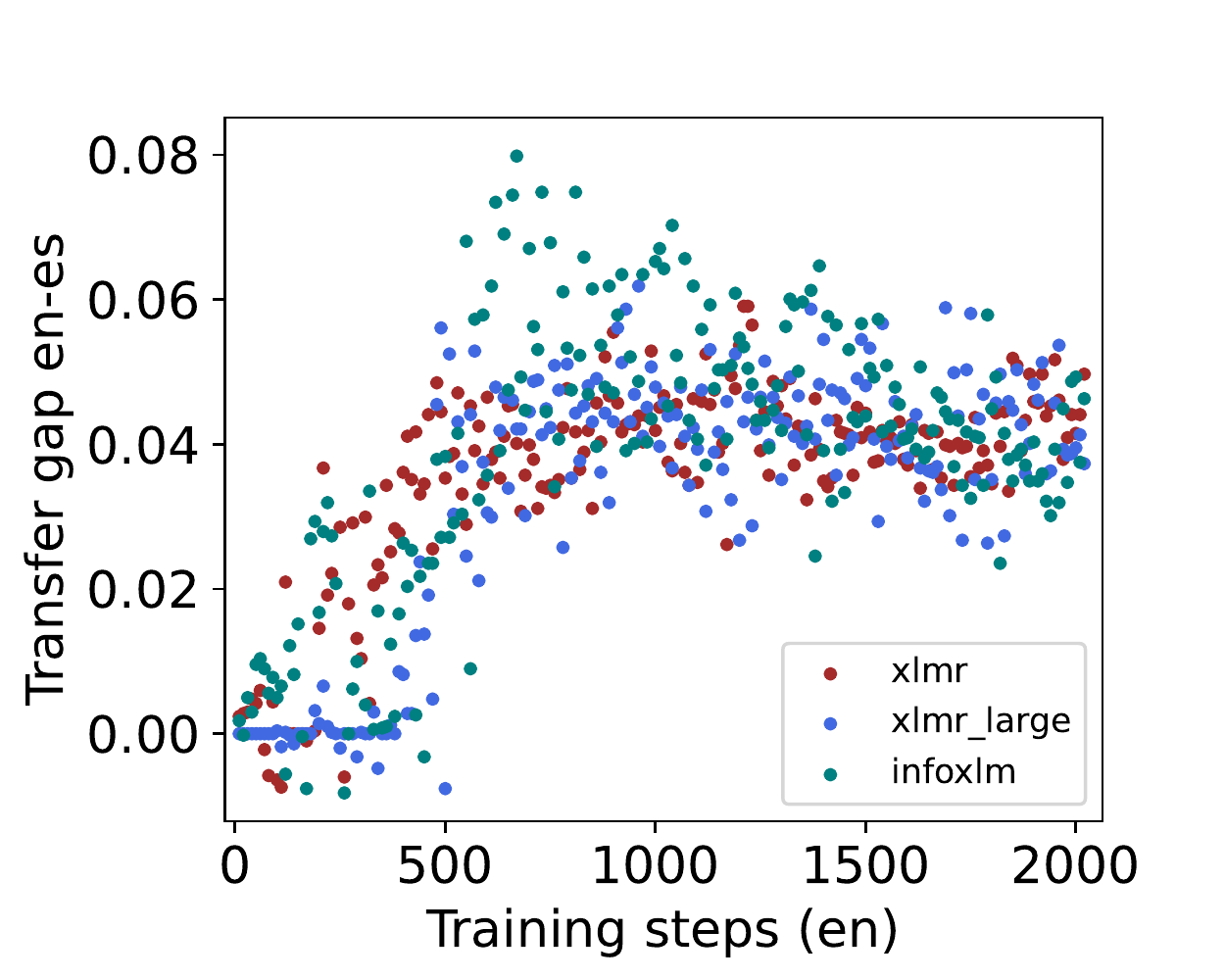}
    \end{subfigure}
    \hfill
    \begin{subfigure}[b]{0.32\textwidth}
     \centering
     \includegraphics[width=0.95\textwidth]{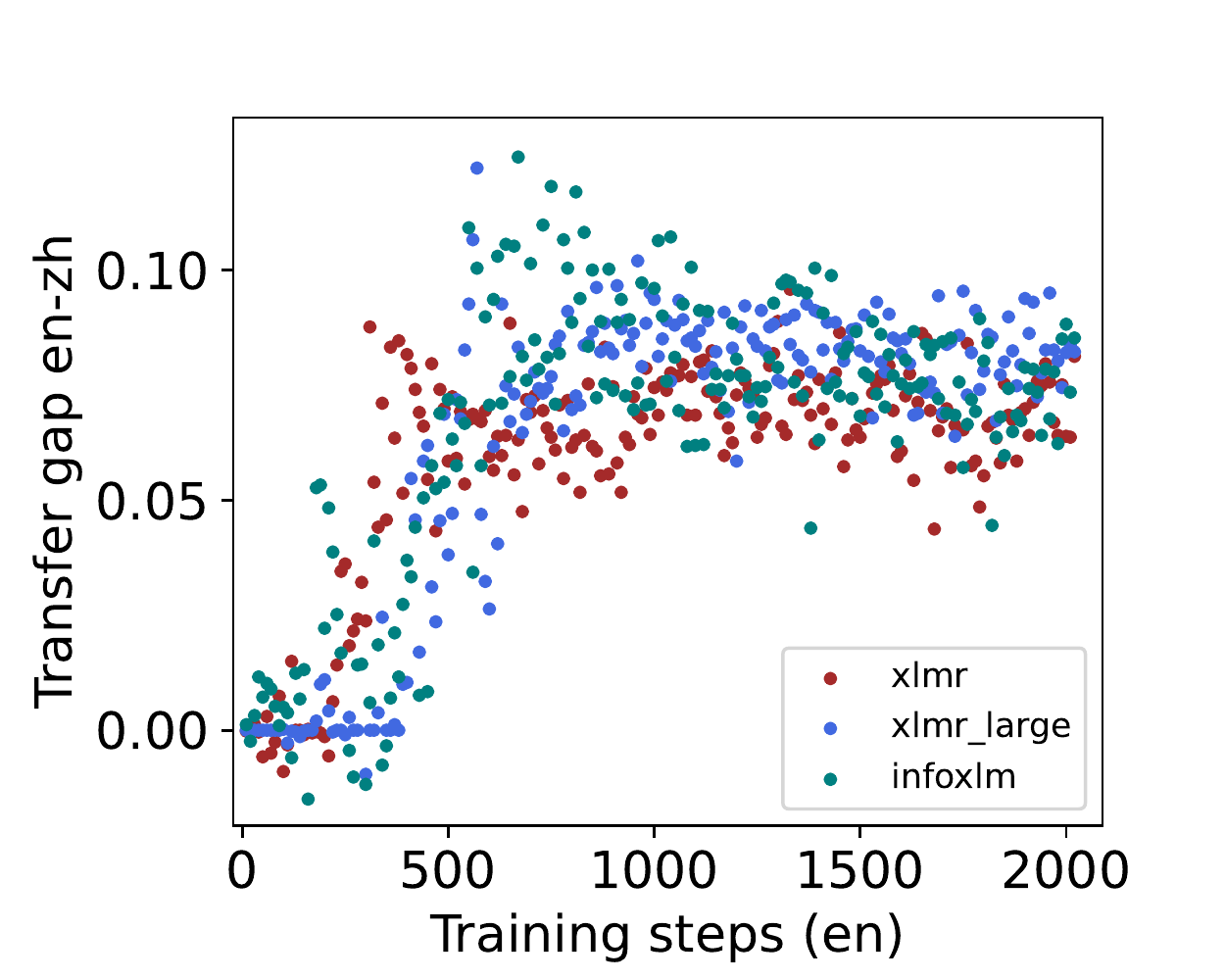}
    \end{subfigure}
    \hfill
    \begin{subfigure}[b]{0.32\textwidth}
     \centering
     \includegraphics[width=0.95\textwidth]{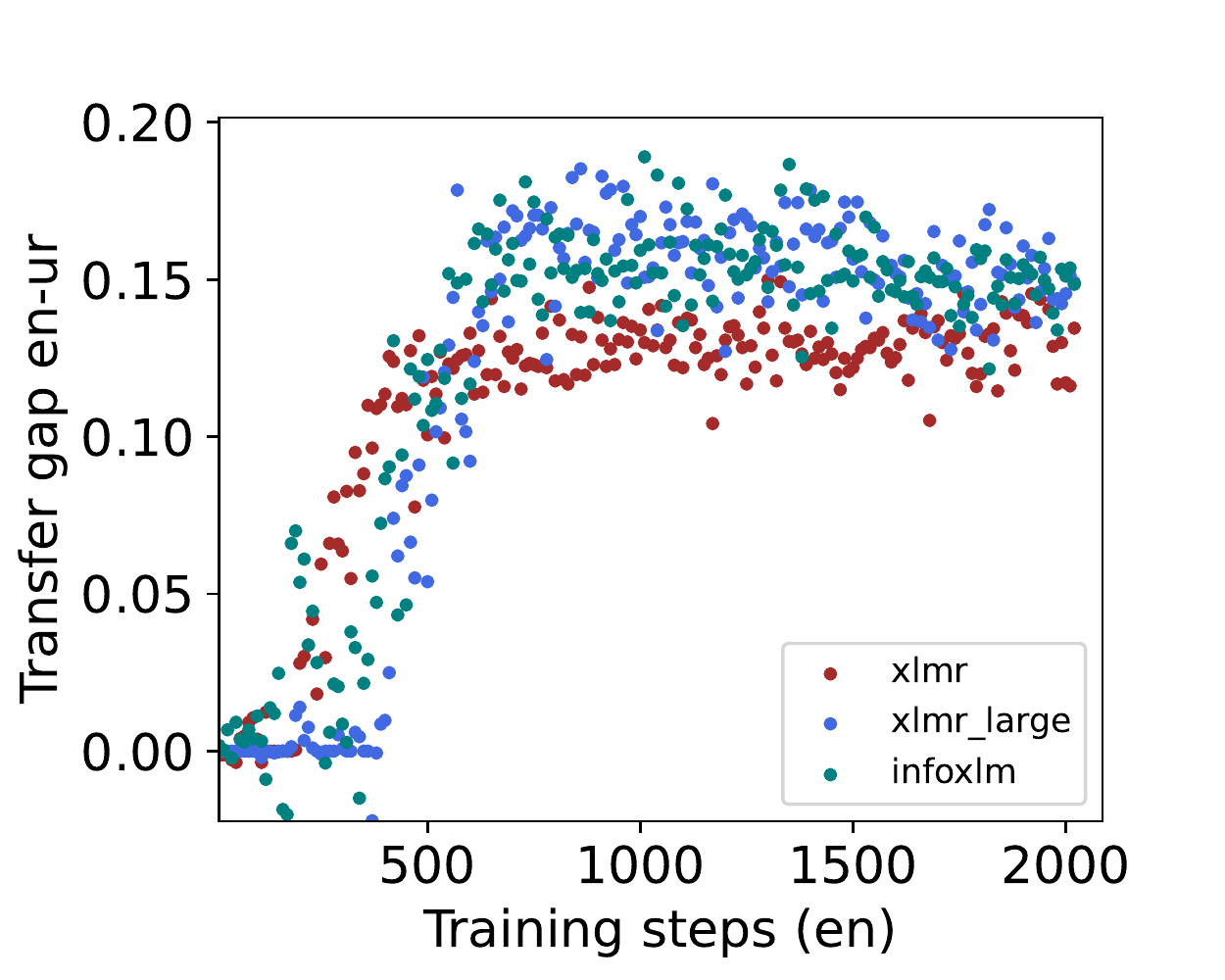}
    \end{subfigure}
\end{subfigure}
\begin{subfigure}[b]{1.0\textwidth}
    \begin{subfigure}[b]{0.32\textwidth}
     \centering
     \includegraphics[width=0.98\textwidth]{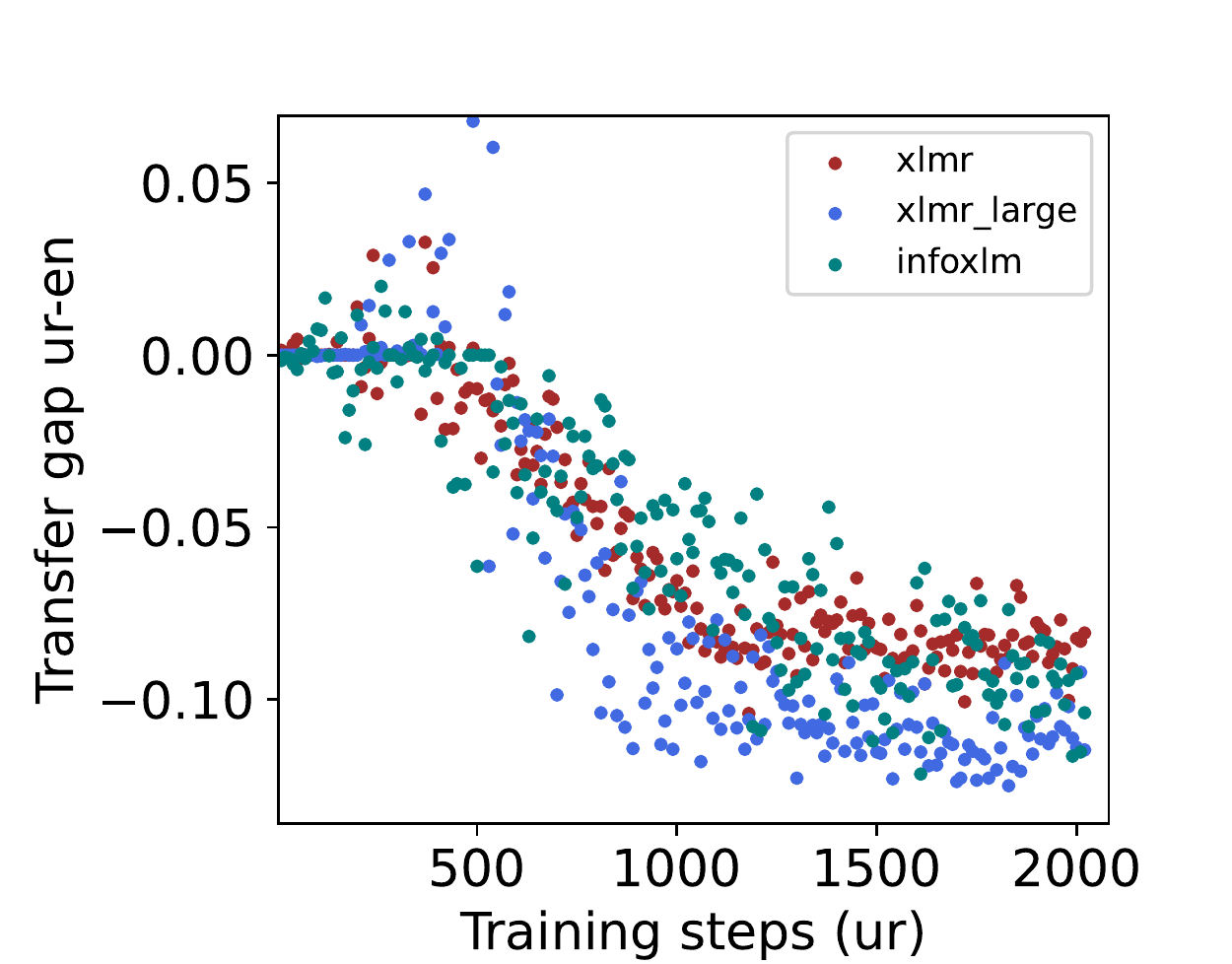}
    \end{subfigure}
    \hfill
    \begin{subfigure}[b]{0.32\textwidth}
     \centering
     \includegraphics[width=0.97\textwidth]{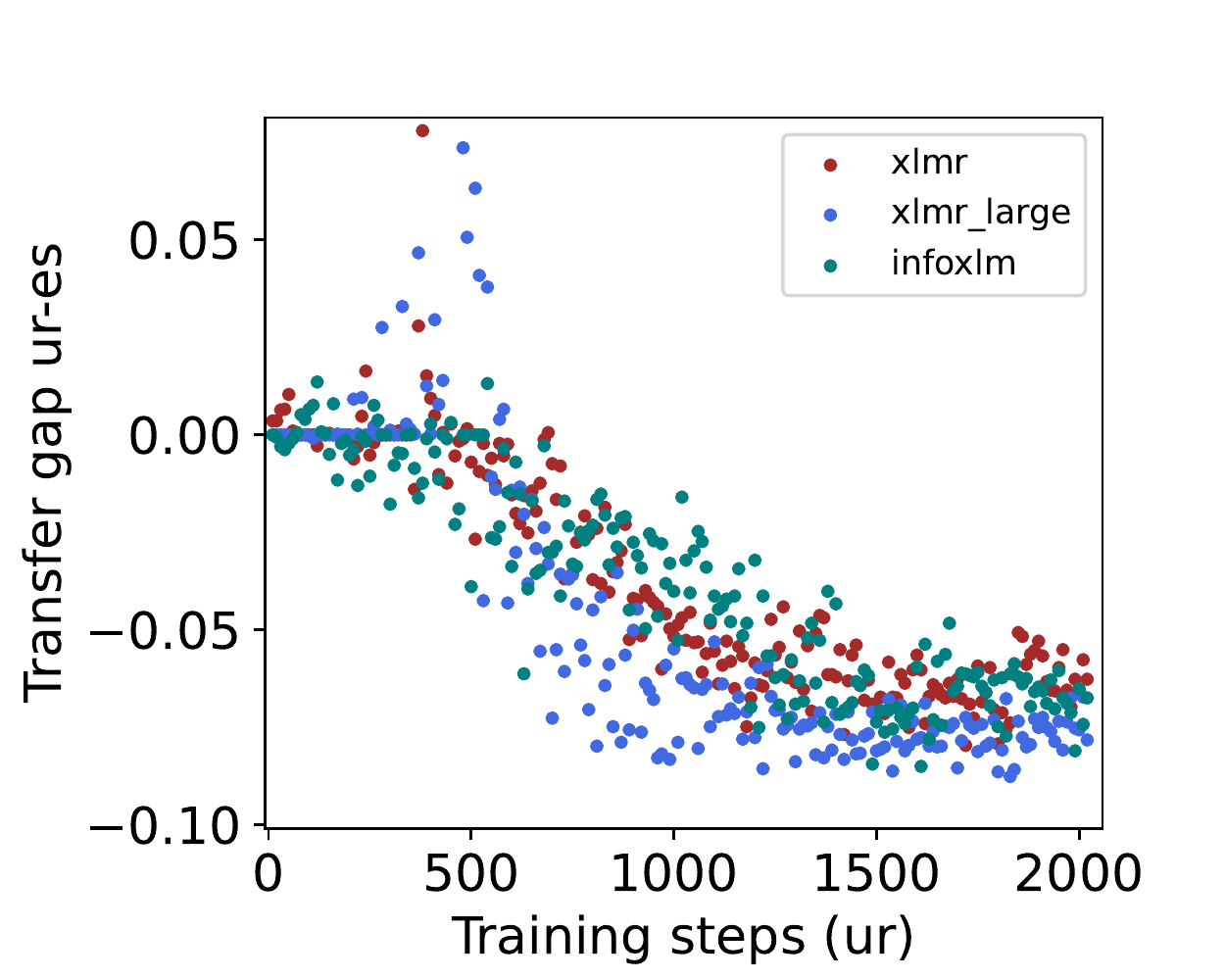}
    \end{subfigure}
    \hfill
    \begin{subfigure}[b]{0.32\textwidth}
     \centering
     \includegraphics[width=0.96\textwidth]{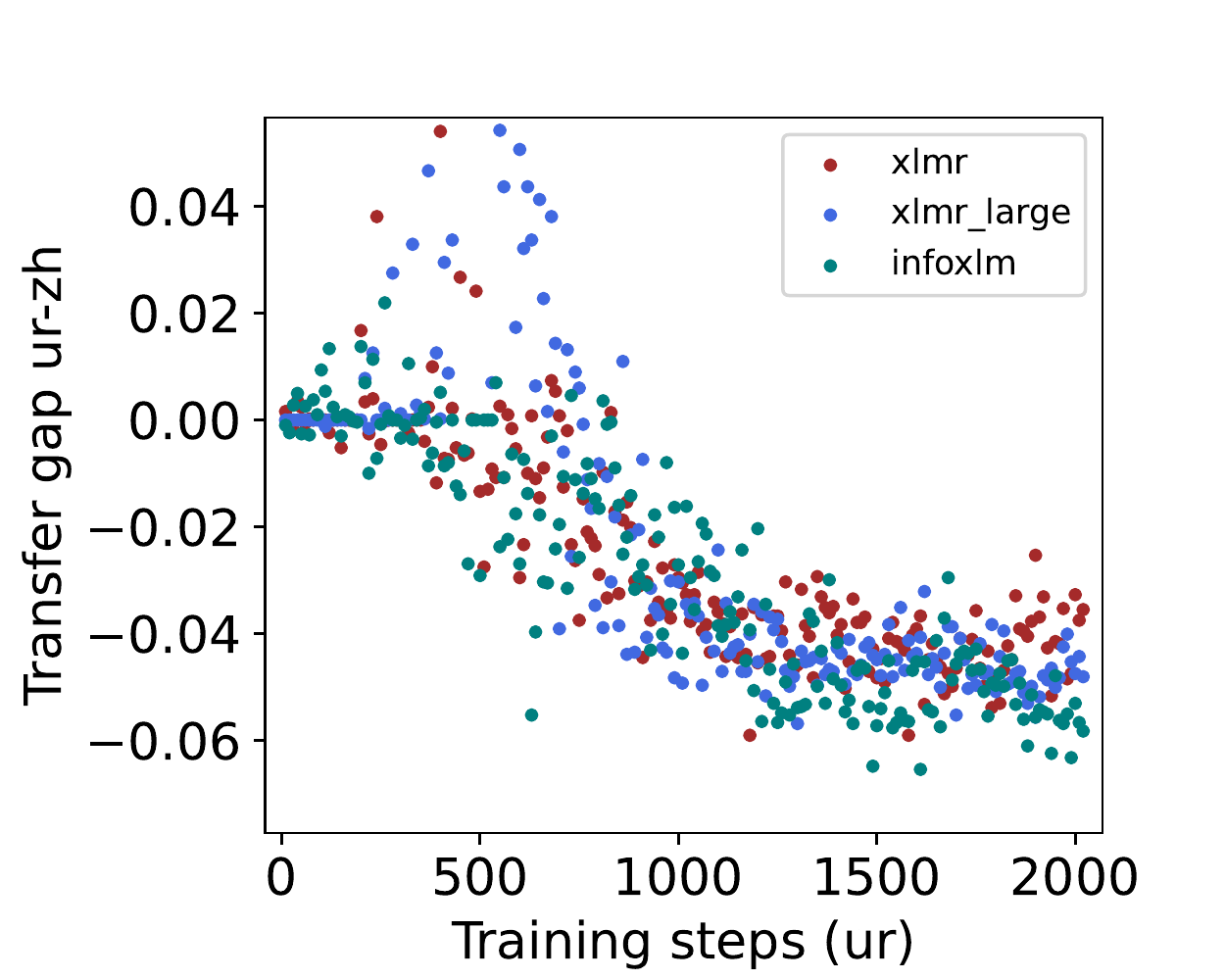}
    \end{subfigure}
    \caption{Cross-lingual transfer gap scores of various multilingual Transformers.}
\end{subfigure}
\begin{subfigure}[b]{1.0\textwidth}
    \begin{subfigure}[b]{0.33\textwidth}
     \centering
     \includegraphics[width=0.95\textwidth]{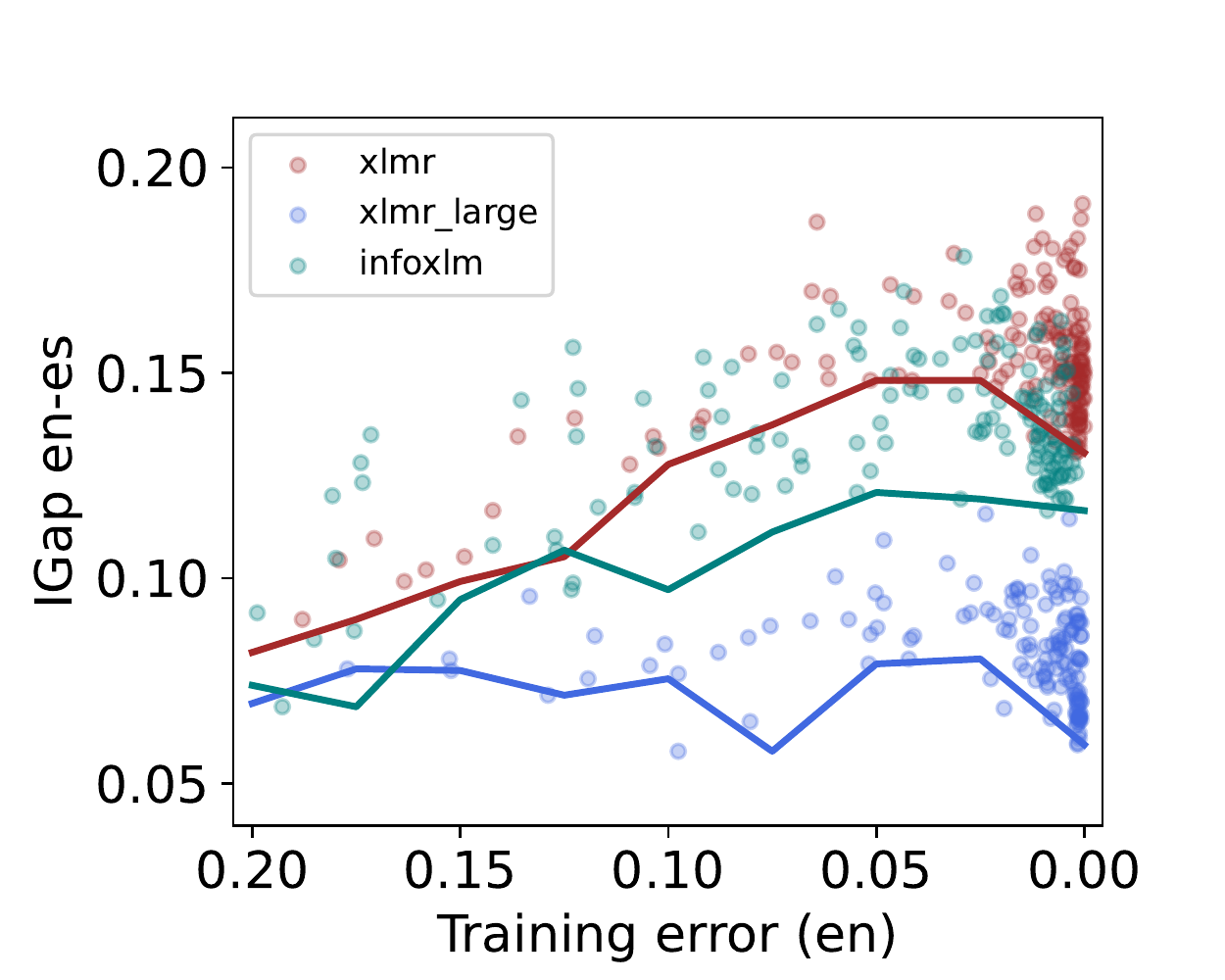}
    \end{subfigure}
    \hfill
    \begin{subfigure}[b]{0.32\textwidth}
     \centering
     \includegraphics[width=0.95\textwidth]{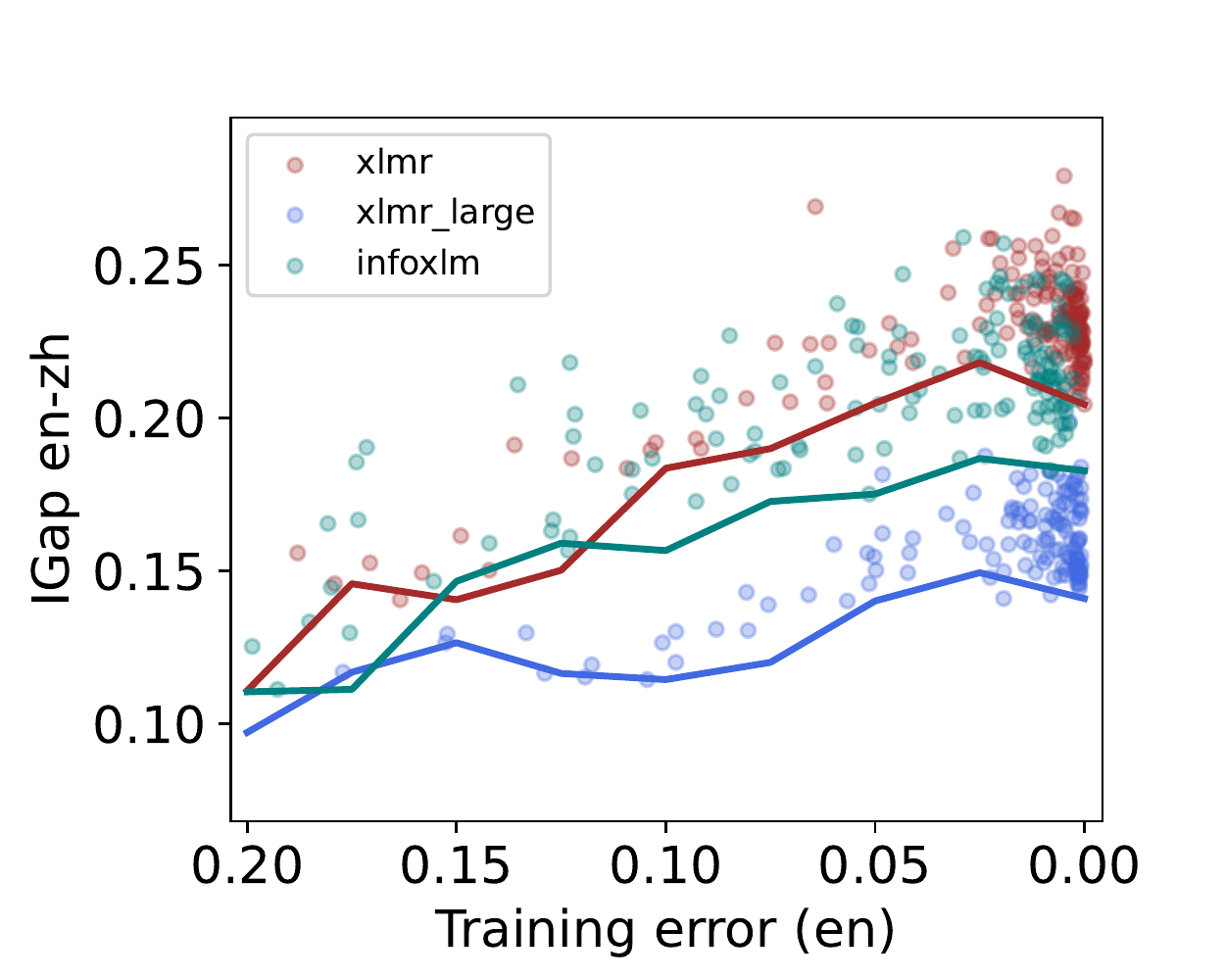}
    \end{subfigure}
    \hfill
    \begin{subfigure}[b]{0.32\textwidth}
     \centering
     \includegraphics[width=0.95\textwidth]{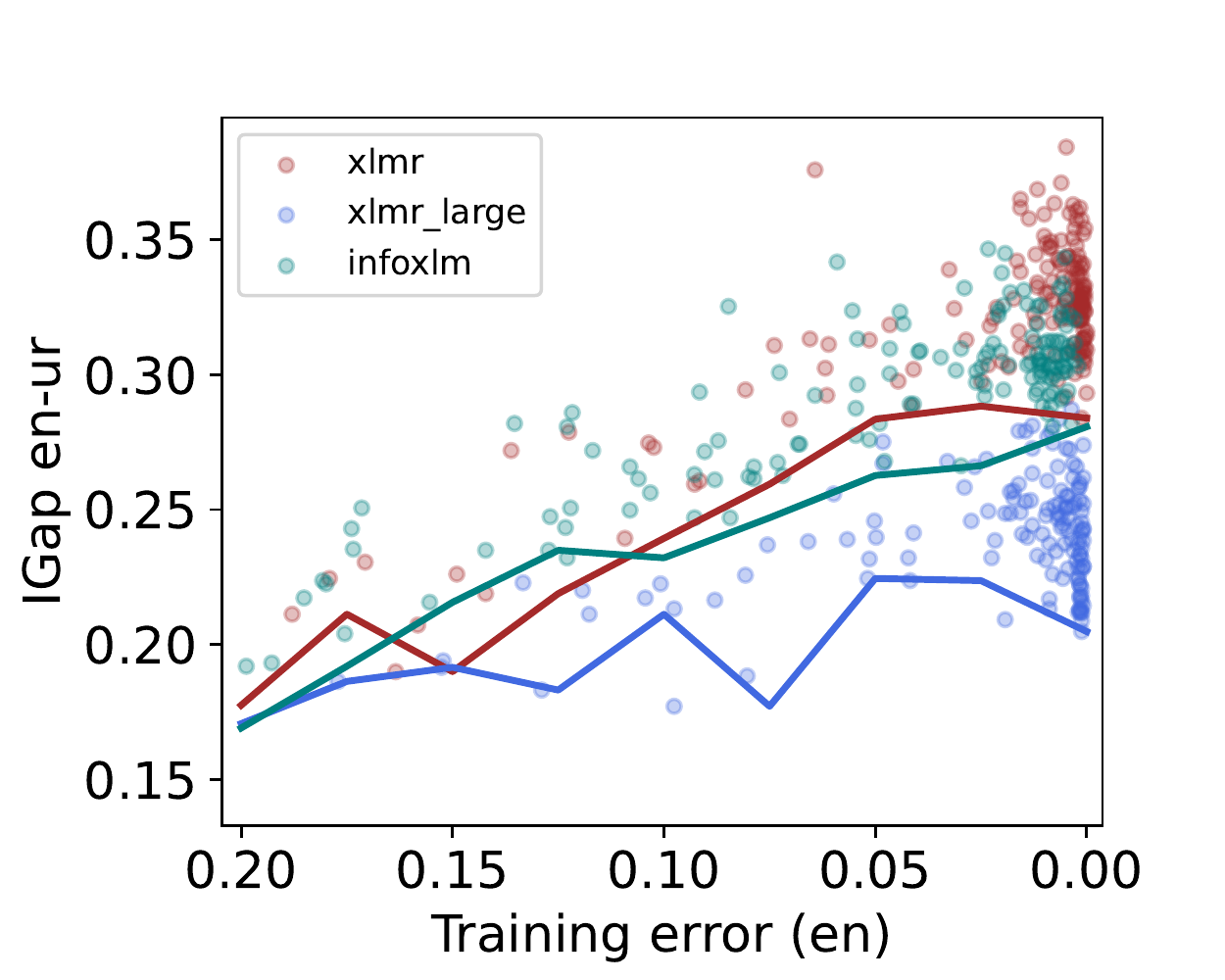}
    \end{subfigure}
\end{subfigure}
\begin{subfigure}[b]{1.0\textwidth}
    \begin{subfigure}[b]{0.32\textwidth}
     \centering
     \includegraphics[width=0.95\textwidth]{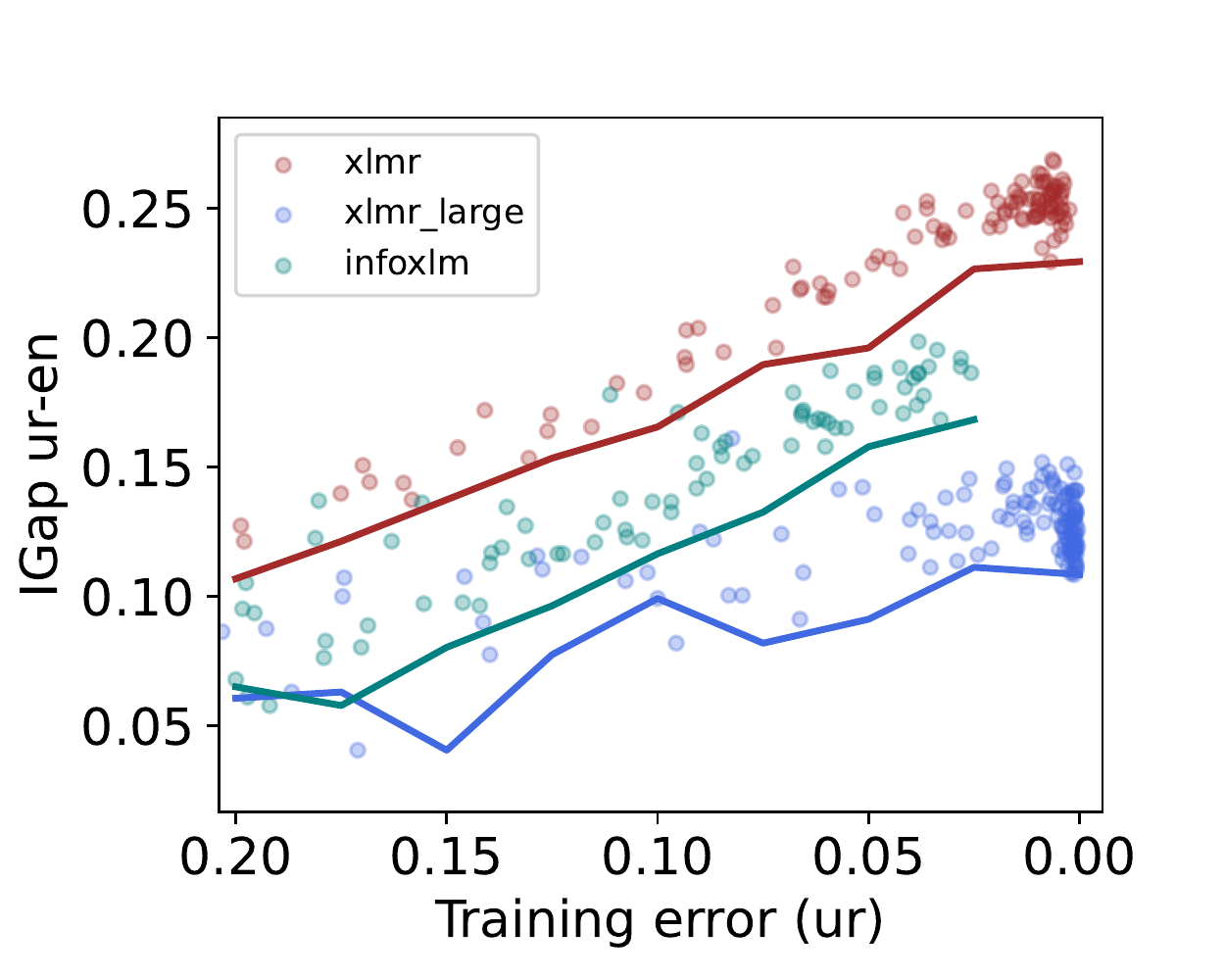}
    \end{subfigure}
    \hfill
    \begin{subfigure}[b]{0.32\textwidth}
     \centering
     \includegraphics[width=0.95\textwidth]{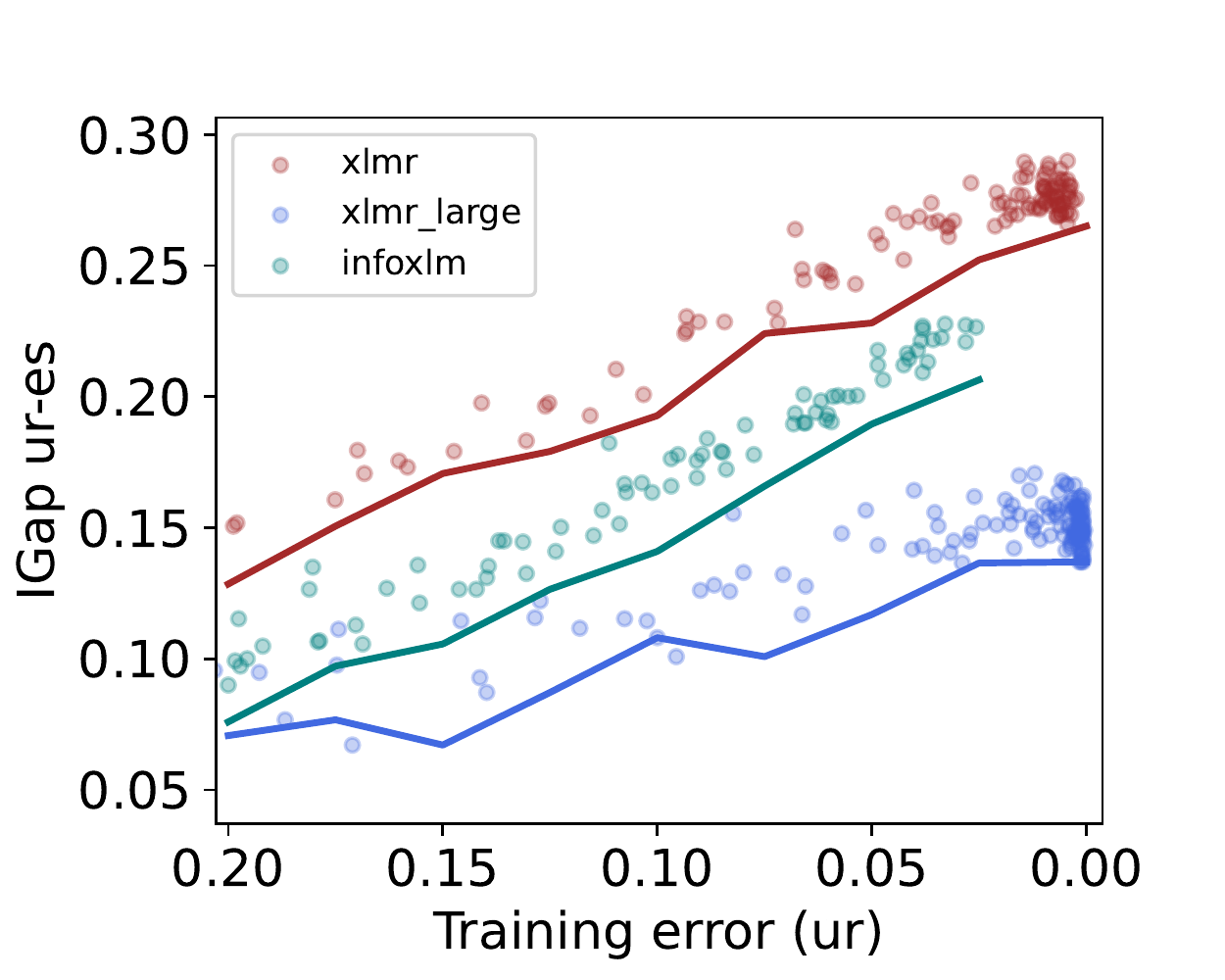}
    \end{subfigure}
    \hfill
    \begin{subfigure}[b]{0.32\textwidth}
     \centering
     \includegraphics[width=0.95\textwidth]{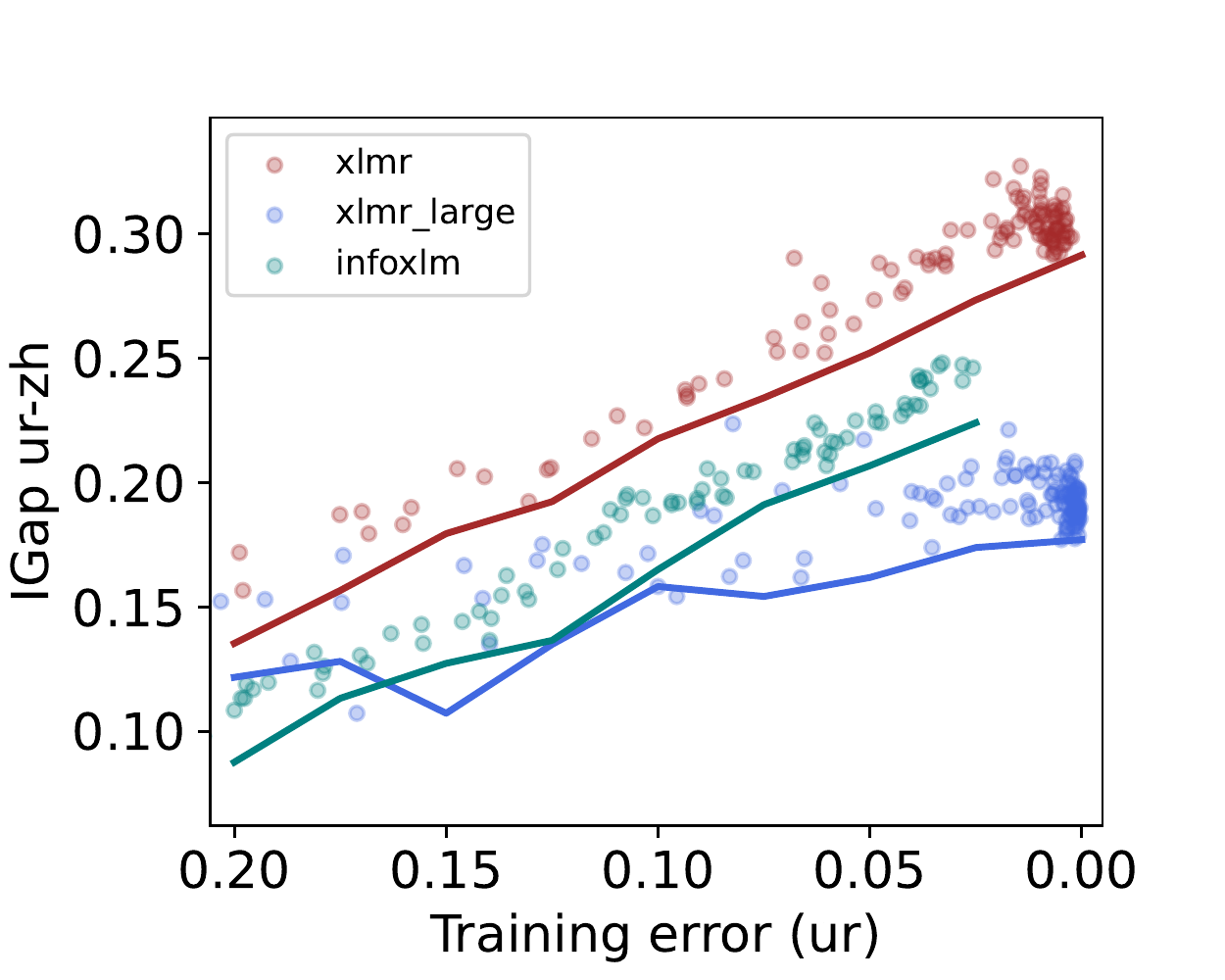}
    \end{subfigure}
    \caption{\our{} curves of various multilingual Transformers.}
\end{subfigure}
\caption{Comparison between \our{} and cross-lingual transfer gap for measuring cross-lingual transferability on XNLI natural language inference, where scores from different models are marked in different colors. }
\label{fig:main}
\end{figure*}

\section{Evaluation}

To validate whether our \our{} metric better indicates cross-lingual transferability, we compare \our{} with cross-lingual transfer gap and representation-based metrics for transferability measuring and transfer direction ranking.

\subsection{Measure transferability}
\label{sec:exp1}

\paragraph{Setup}
We measure \our{} and transfer gap scores of the following three multilingual Transformer models. (1) XLM-R~\citep{xlmr} is pre-trained on large-scale multilingual text corpora with the masked language modeling task. (2) XLM-R$_\text{large}$ is the large-size version of XLM-R with more parameters. (3) InfoXLM~\citep{infoxlm} enhances cross-lingual abilities by learning the cross-lingual contrast task on parallel data. The end task is natural language inference (NLI), intending to classify the input text pair into three categories. We use the validation sets of XNLI~\citep{xnli} to compute the metrics,
which provides validation and text examples in English, and human-translated examples in other languages. 

\paragraph{Results}
Figure~\ref{fig:main} compares \our{} with cross-lingual transfer gap for measuring transferability on XNLI, where we compute the metric scores using fine-tuned models with various training steps with a step size of $10$. In Figure~\ref{fig:main}a, each point represents the transfer gap score of a fine-tuned model. In Figure~\ref{fig:main}b, the points stand for $\mathcal{G}_\text{inter}$ scores, and we compute the \our{} curves using Equation~(\ref{eq:igap}) with $\epsilon = 0.025$ and training errors $\mathcal{E}'$ ranging from $0.2$ to $0$ with a step size of $0.025$.

Overall, \our{} shows clear curves and consistent results among six different transfer directions. Most of the time, XLM-R, InfoXLM, and XLM-R$_\text{large}$ are in descending order of \our{}. On the contrary, Figure~\ref{fig:main} (a) shows that the transfer gap scores of the three models are mixed up and inconsistent across transfer directions. In Figure~\ref{fig:main}a, XLM-R, InfoXLM, and XLM-R$_\text{large}$ achieve the lowest transfer gap in the transfer direction of en-ur, ur-en, and ur-zh, respectively. Moreover, cross-lingual transfer gap can either increase or decrease when the models are trained with more step, and shows negative transferability. In contrast, \our{} shows not only consistent positive transferability but also a consistent increasing trend when models have lower training errors. In summary, the clear curves and consistent results demonstrate that \our{} indicates cross-lingual transferability better than cross-lingual transfer gap.

\begin{table*}[t]
\caption{Transfer direction ranking accuracy on XNLI. We compare \our{} with three representation-based metrics, i.e., $L_2$ distance, dot product, and cosine similarity. Cross-lingual transfer gap is not included because it relies on end-task data. The models are fine-tuned with three random seeds.}
\centering
\scalebox{0.85}{
\renewcommand\tabcolsep{4.5pt}
\begin{tabular}{lcccccccccccccccc}
\toprule
\bf Metric &  ar  &   bg  &   de  &   el  &   en  &   es  &   fr  &   hi  &   ru  &   sw  &   th  &   tr  &   ur  &   vi  &   zh  &  avg \\ \midrule
$L_2$ & 57.1 &   73.6 &   80.2 &   71.4 &   75.8 &   79.1 &   78.0 &   47.3 &   83.5 & \bf  72.5 &   37.4 &   60.4 &   28.6 & \bf  84.6 &   50.5 &   65.3 \\
\textsc{Dot} & 47.3 &   63.7 &   68.1 &   61.5 &   67.0 &   64.8 &   67.0 &   47.3 &   71.4 &   56.0 &   45.1 &   56.0 &   45.1 &   64.8 &   49.5 &   58.3 \\
\textsc{Cos} & 53.8 &   73.6 &   79.1 &   69.2 &   78.0 &   79.1 &   78.0 &   47.3 &   83.5 &   71.4 &   37.4 &   61.5 &   28.6 & \bf  84.6 &   49.5 &   65.0 \\
\our{} & \bf 70.3 &  \bf 75.8 &  \bf 83.5 & \bf  79.1 & \bf  84.6 & \bf  86.8 & \bf  85.7 & \bf  60.4 & \bf  92.3 &   70.3 & \bf  82.4 &  \bf 82.4 &  \bf 46.2 &   54.9 & \bf  76.9 & \bf  75.5 \\
\bottomrule
\end{tabular}
}
\label{table:tdr}
\end{table*}

\subsection{Transfer direction ranking}
\label{sec:tdr}

\paragraph{Task description}
In addition to transferability measuring, we present a task to evaluate cross-lingual transferability metrics without end-task data, called \textit{transfer direction ranking} (TDR). For a specific source language, the goal is to predict a target language sequence without end-task data, in which the languages are sorted by the cross-lingual transferability. The golden sequence is obtained by performing cross-lingual transfer on the end task. The predicted sequence is evaluated by comparing the predicted sequence with the golden sequence. Formally, let $S = \{ l_1, l_2, \dots, l_n \}$ denote the golden language sequence, where the model achieves the best transfer performance in target language $l_1$. Similarly, let $\hat{S} = \{ \hat{l}_1, \hat{l}_2, \dots, \hat{l}_n \}$ denote the sequence predicted by \our{}. The TDR accuracy is computed by
\begin{align}
\begin{split}
    \text{acc} = \frac{2}{n(n-1)} \sum_{l_1', l_2'} \mathbbm{1} \{ (I_{S}(l_1') - I_{S}(l_2')) \times (I_{\hat{S}}(l_1') - I_{\hat{S}}(l_2')) > 0 \},
\end{split}
\end{align}
where $(l_1', l_2')$ means all unique $2$-combination language pairs, and $I_{S}(l_1')$ means the index of $l_1'$ in the sequence $S$.

\paragraph{Setup} We evaluate fine-tuned XLM-R models on XNLI validation sets in all $15 \times 15$ transfer directions to obtain the gold target language sequences for all the source languages. Then, we predict the target language sequence by computing \our{} without XNLI data but parallel sentences from the development set of FLORES-101~\citep{flores101}. We estimate transferability with randomly-generated labels as mentioned in Section~\ref{sec:igap}. We compare \our{} with three representation-based metrics for transfer direction ranking, including $L_2$ distance, dot-product, and cosine similarity. 
We estimate the transferability by representation similarities, because the aligned representations are typically considered one of the elements of transferability.
Following~\citet{xtreme}, we utilize the average hidden vectors as the sentence representation, and compute the similarities by measuring the above three metrics over parallel sentences from the development set of FLORES-101. The hidden vectors are from the $7$-th layer because the layer has the best-aligned representations \citep{infoxlm}. Cross-lingual transfer gap is not included because it relies on end-task data.

\paragraph{Results}
Table~\ref{table:tdr} presents the evaluation results, where each number means the TDR accuracy for the transfer directions from a specific source language to various target languages. \our{} achieves the best TDR accuracy in $13$ out of $15$ languages, demonstrating the effectiveness of \our{} for the estimation of cross-lingual transferability without end-task data. In contrast, representation-based metrics perform less well than \our{}. For a specific source language, a target language may have more similar representations than another target language, but it can still perform worse for cross-lingual transfer. It may seem counterintuitive, but the results demonstrate that better-aligned representations do not promise better cross-lingual transferability or transfer results. 

\begin{table}[t!]
\caption{Average \our{} and test accuracy over target languages of multilingual Transformers.}
\centering
\scalebox{0.87}{
\begin{tabular}{l|cccc}
\toprule
\multirow{2}{*}{\bf Model} & \multicolumn{2}{c}{XNLI} & \multicolumn{2}{c}{PAWS-X} \\
 & test acc$\uparrow$ & \our{}$\downarrow$ & test acc$\uparrow$ & \our{}$\downarrow$ \\ \midrule
 mBERT & 59.5 & 34.9 & 74.1 & 21.6 \\
 XLM & 62.9 & 27.6 & 75.4 & 23.3 \\
 XLM-R & 65.7 & 21.1 & 80.1 & 16.8 \\
 InfoXLM & 66.6 & 20.0 & 82.9 & 14.0 \\
 XLM-R$_\text{large}$ & \bf 74.4 & \bf 13.1 & \bf 85.1 & \bf 12.9 \\
\bottomrule
\end{tabular}
}
\label{table:models}
\end{table}

\begin{table*}[t]
\caption{\our{} scores for various fine-tuning algorithms for cross-lingual transfer. We fine-tune XLM-R using three different fine-tuning algorithms with three random seeds, and compute \our{} scores from English to the other $14$ languages. The last column shows the average XNLI accuracy over the test sets of the $14$ languages.}
\centering
\scalebox{0.83}{
\renewcommand\tabcolsep{4.0pt}
\begin{tabular}{l|ccccccccccccccc|c}
\toprule
\bf Algorithm &  ar  &   bg  &   de  &   el   &   es  &   fr  &   hi  &   ru  &   sw  &   th  &   tr  &   ur  &   vi  &   zh  &  avg & test acc \\ \midrule
\textsc{Vanilla} & 23.1 &  15.9 &  15.6 &  18.1 &  11.6 &  15.9 &  26.0 &  20.7 &  31.9 &  23.1 &  23.7 &  30.1 &  18.6 &  21.1 &  21.1 & 65.7 \\
\textsc{Gaussian} & 21.8 & \bf 14.7 & \bf 13.2 &  16.6 & \bf  10.5 & \bf 13.9 & 
 \bf 23.9 &  19.7 &  30.9 &  21.4 & \bf 21.2 &  28.9 & \bf 16.6 & \bf 19.3 &  \bf 19.5 & 66.1 \\
\textsc{xTune} & \bf 21.7 &  15.3 &  13.4 & \bf 15.8 &  11.0 & \bf 13.9 &  24.3 & \bf 19.1 & \bf 30.6 & \bf 21.0 &  22.1 & \bf 28.4 &  17.0 &  19.7 & \bf 19.5 & \bf 67.1 \\
\bottomrule
\end{tabular}
}
\label{table:algo}
\end{table*}

\section{Empirical analyses}

\subsection{Compare multilingual transformers}
\label{sec:compare-model}
We explore how multilingual Transformers differ in terms of cross-lingual transferability. Specifically, we compare \our{} scores of five widely-used multilingual Transformers. In addition to the three models mentioned in Section~\ref{sec:exp1}, we also include the following two models.
(1) mBERT~\citep{bert} is a multilingual version of BERT. (2) XLM~\citep{xlm} enhances cross-lingual pre-training with translation language modeling. We compute the \our{} scores of the models with $\epsilon = 0.001$ and training error $\mathcal{E}' = 0$ using English as the source language. The end tasks include natural language inference on XNLI~\citep{xnli} and text classification on PAWS-X~\citep{pawsx}. Both datasets provide human-translated validation sets in multiple languages.

Table~\ref{table:models} compares the test accuracy and \our{} scores averaged over the target languages. The evaluated multilingual Transformer models obtain various accuracy and \our{} scores, showing that the models have different cross-lingual transferability. 
Compared to mBERT, the other models typically achieve better test accuracy and reduce \our{}. Detailed results can be found in Appendix~\ref{sec:app-results}.

\subsection{Compare fine-tuning algorithms}
\label{sec:compare-algo}
We explore how fine-tuning algorithms influence cross-lingual transferability. We fine-tune XLM-R on XNLI using the following three fine-tuning algorithms under the zero-shot transfer setting with English as the source language. (1) \textsc{Vanilla} directly fine-tunes the model on the training data in the source language. (2) \textsc{Gaussian} follows \citet{xquad}, which adds Gaussian noise to the word embeddings during fine-tuning. (3) \textsc{xTune}~\citep{xtune} employs an consistency regularization loss. We implement a stage-1 \textsc{xTune} with Gaussian-noise data augmentation, which empirically performs well. We perform fine-tuning with the three algorithms on the validation set of XNLI and PAWS-X and compute the \our{} scores with $\epsilon = 0.001$ and training error $\mathcal{E}' = 0$.

Table~\ref{table:algo} and Table~\ref{table:app-ca} compare the \our{} scores across fine-tuning algorithms on XNLI and PAWS-X, and the last columns show the test accuracy averaged over target languages. Both \textsc{Gaussian} and \textsc{xTune} improve the cross-lingual transfer performance and \textsc{xTune} achieves the best results, which is consistent with the results reported by \citet{xtune}. Correspondingly, both \textsc{Gaussian} and \textsc{xTune} consistently have lower \our{} than the \textsc{Vanilla} fine-tuning algorithm. The results suggest that fine-tuning algorithms for cross-lingual transfer implicitly reduce \our{}, i.e., having better cross-lingual transferability.

\begin{figure*}[t]
\hfill
\begin{subfigure}[b]{0.32\textwidth}
 \centering
 \includegraphics[width=0.98\textwidth]{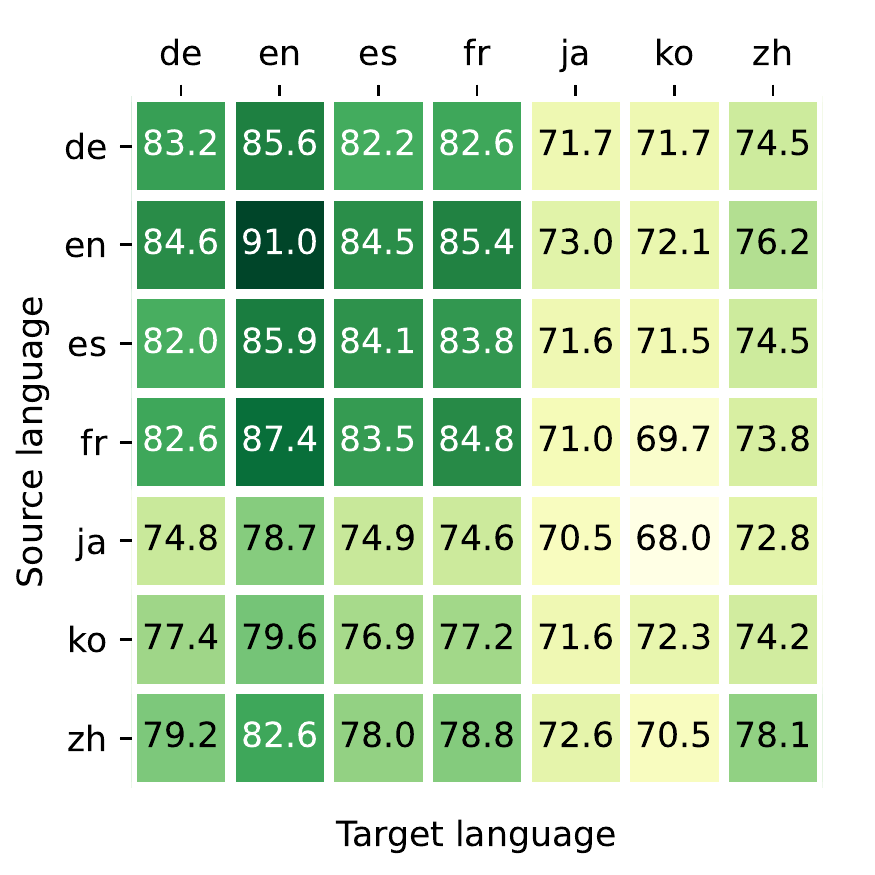}
 \caption{Test accuracy.}
 \label{fig:td1}
\end{subfigure}
\hfill
\begin{subfigure}[b]{0.32\textwidth}
 \centering
 \includegraphics[width=0.98\textwidth]{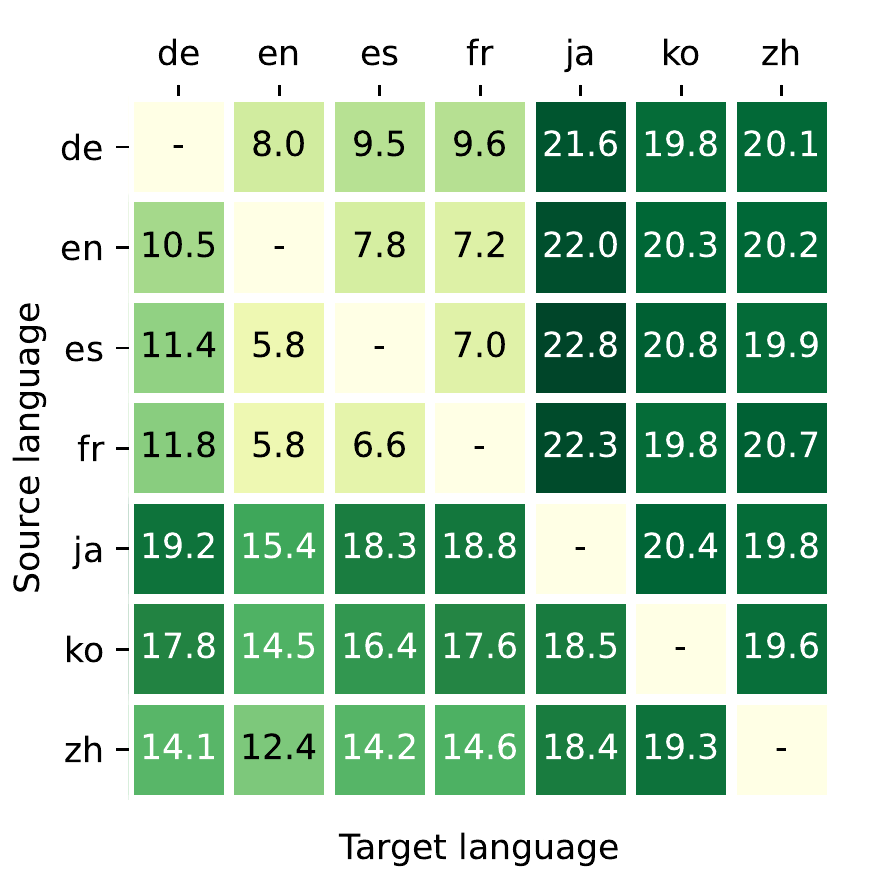}
 \caption{\our{}.}
 \label{fig:td2}
\end{subfigure}
\hfill
\begin{subfigure}[b]{0.32\textwidth}
 \centering
 \includegraphics[width=0.98\textwidth]{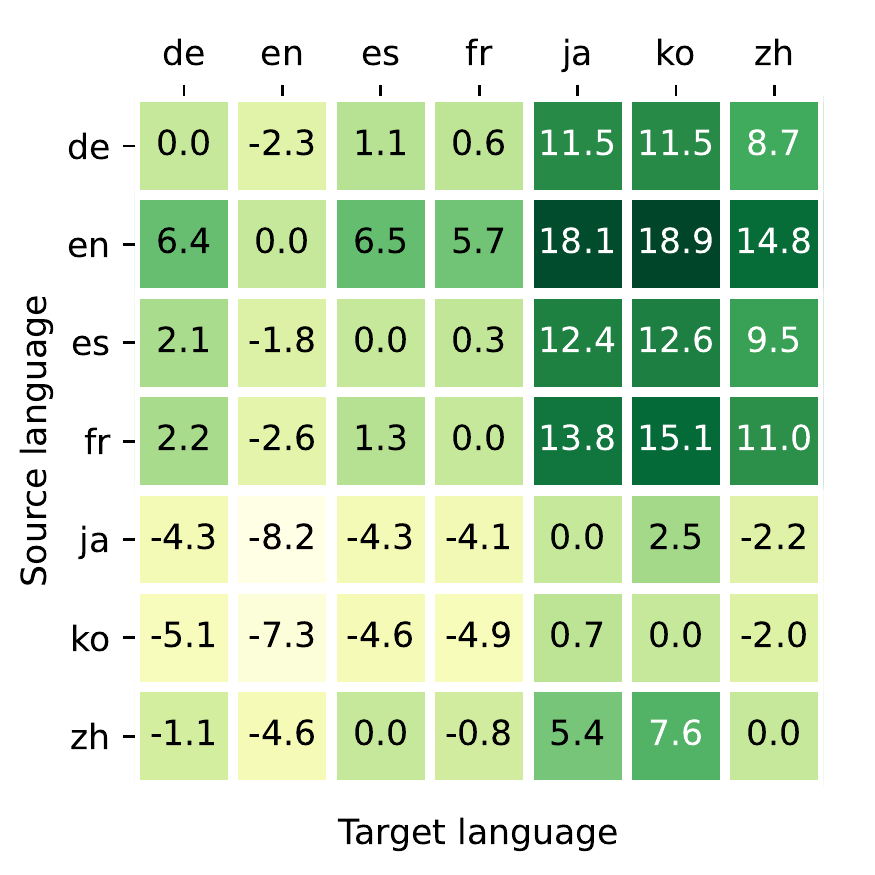}
 \caption{Cross-lingual transfer gap.}
 \label{fig:td3}
\end{subfigure}
\hfill
\caption{Comparison of test accuracy, \our{}, and transfer gap among $7\times7$ transfer directions on PAWS-X.}
\label{fig:td}
\end{figure*}

\subsection{Compare transfer directions}
\label{sec:ctd}

We investigate how transfer direction affects cross-lingual transferability. For each target language, we fine-tune XLM-R on PAWS-X and XNLI with each language as source language separately. After fine-tuning with three random seeds, we present the transfer performance, \our{}, and transfer gap in Figure~\ref{fig:td}. 
See Appendix~\ref{sec:app-results} for the results on XNLI. It can be observed that for a specific target language, using different source languages can lead to different transfer results and \our{} scores. Besides, we find that the \our{} matrix is asymmetric, indicating that the transferability can be different when reversing the transfer direction. The results also show that cross-lingual transfer gap not only produces negative gap scores, but also fails to measure transferability. For instance, in Figure~\ref{fig:td2} and Figure~\ref{fig:td3}, the column of ``fr'' show the \our{} and transfer gap scores from source languages to English. We can observe that \our{} successfully indicates that English is the best source language. In contrast, transfer gap indicates Korean has the lowest gap, which performs less well as shown in Figure~\ref{fig:td1}.

\begin{figure}[t]
\hspace*{\fill}
\begin{subfigure}[b]{0.33\textwidth}
 \centering
 \includegraphics[width=0.95\textwidth]{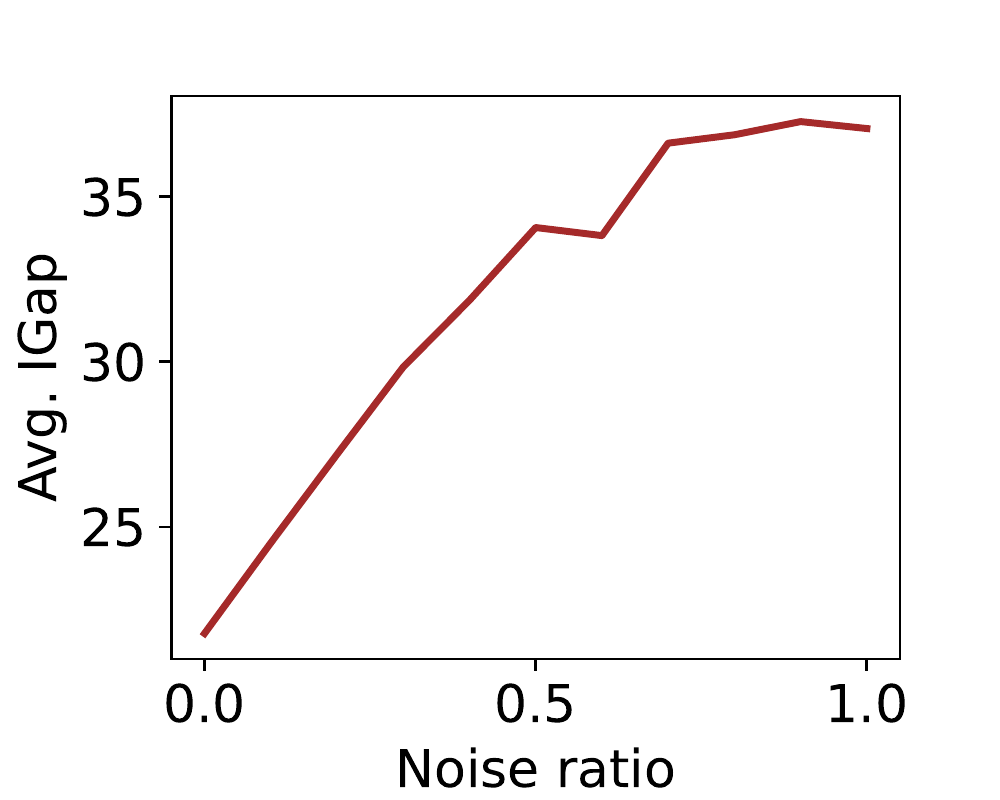}
 \caption{\our{}-noise curve.}
 \label{fig:mem1}
\end{subfigure}
\hfill
\begin{subfigure}[b]{0.33\textwidth}
 \centering
 \includegraphics[width=0.95\textwidth]{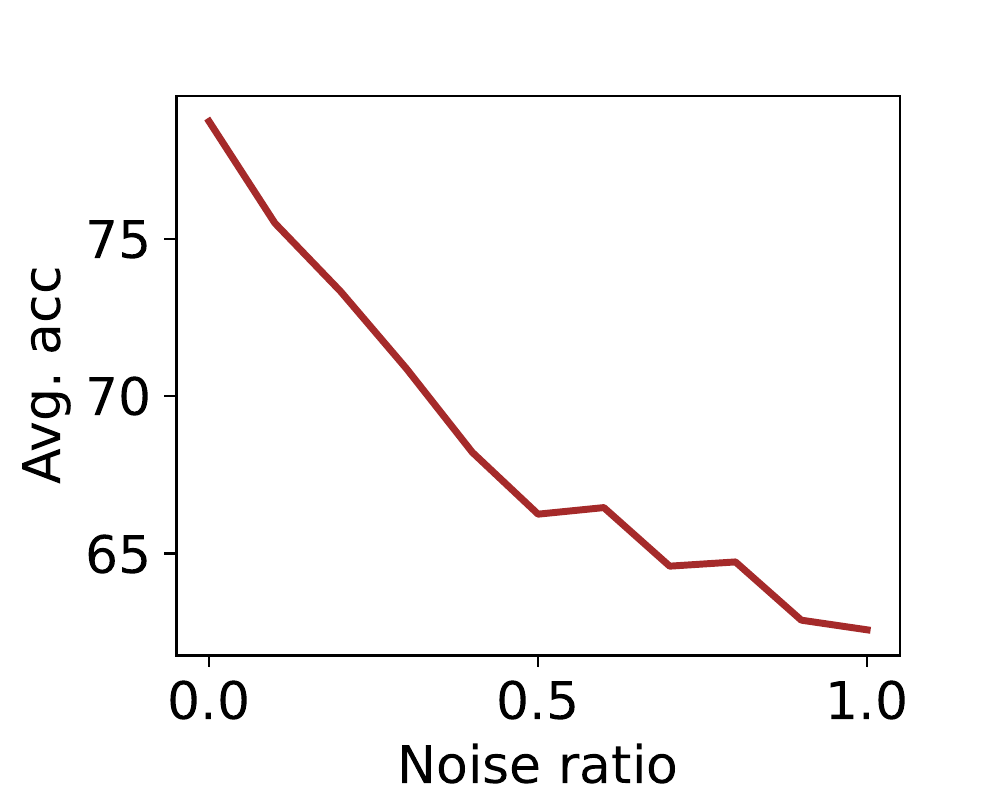}
 \caption{Acc-noise curve.}
 \label{fig:mem2}
\end{subfigure}
\hspace*{\fill}
\caption{Memorization effects of multilingual Transformers. We fine-tune XLM-R on partially corrupted XNLI with randomly-generated labels, and then predict the labels in other languages.}
\label{fig:mem}
\end{figure}

\subsection{Memorization effect}
\label{sec:mem}
To better understand the cross-lingual transferability, we conduct experiments on the memorization effects using \our{} as a tool. Our goal is to investigate whether cross-lingual transfer can transfer newly-learned knowledge across languages. Inspired by \citet{zhang2017understanding}, we extend the randomization tests to the cross-lingual transfer setting. The original randomization tests are designed to understand the effective capacities of neural networks, but here we do not focus on model capacity but on whether the newly-memorized information can be transferred to other languages. Following \citet{zhang2017understanding}, we construct corrupted training sets by partially adding noise by assigning examples with randomly-generated labels. Then, we fine-tune XLM-R on the corrupted datasets, and study whether the randomly-generated labels can be predicted in other languages.

Figure~\ref{fig:mem} illustrates the average classification accuracy and average \our{} scores of the target languages, where the noise ratio ranges from $0$ to $1.0$ with a step size of $0.1$. We obtain lower classification accuracy and higher \our{} when enlarging the corruption ratio, showing that the randomly-generated labels are more difficult to transfer than the end-task knowledge. Moreover, the model performs surprisingly well on target languages even with a corruption ratio of $1.0$, i.e., all labels are randomly generated. Empirically, if the randomly-generated labels are not transferred, the accuracy and \our{} should be $33.3\%$ and $66.7\%$ for random prediction, respectively. Differently, the models achieve accuracy and \our{} of $62.6$ and $37.0$, respectively, demonstrating that even randomly-generated knowledge can also be transferred.

\section{Discussion}
Through extensive systematic experiments, we have the following findings.

\paragraph{Cross-lingual transfer methods implicitly reduce \our{}.} Our experiments in Section~\ref{sec:compare-model} and Section~\ref{sec:compare-algo} compare \our{} among multilingual Transformers and fine-tuning algorithms. Although they are designed to achieve better cross-lingual transfer performance, our experimental results demonstrate that they reduce \our{} implicitly, i.e., having better cross-lingual transferability.

\paragraph{Multilingual Transformers can memorize new knowledge and transfer it to other languages.} In Section~\ref{sec:mem}, we assign text with randomly-generated labels as new knowledge to be transferred. Using \our{}, we measure the transferability on corrupted datasets. We show that multilingual Transformers memorize randomly-generated labels and can predict the labels in other languages.

\paragraph{Better-aligned representations do not promise better cross-lingual transferability.} In Section~\ref{sec:tdr}, we show that multilingual Transformers may learn well-aligned representations between two languages, but it does not indicate good transferability from one language to another. In Table~\ref{table:tdr}, representation similarity metrics only successfully predict about $65\%$ of the language orders, demonstrating that better-aligned representations do not promise better cross-lingual transferability.

\section{Related work}

Starting from Multilingual BERT (mBERT;~\citealt{bert}), multilingual Transformers have been developed as the backbone for a wide range of NLP tasks \citep{xlmr,mt5,mbart,veco}. The models are further improved in terms of representation alignment \citep{labse,hictl,hu2020explicit,xlmalign}, training scales \citep{mt5,scao2022bloom,palm,chung2023unimax}, and cross-lingual transferability \citep{xlm,erniem,infoxlm}.

The capabilities of multilingual Transformers have been explored in many aspects. \citet{wu2019beto} and \citet{pires2019multilingual} show that mBERT achieves zero-shot cross-lingual transfer because of its cross-lingual transferability.
The transferability is also observed from other multilingual Transformers \citep{xlm,xtreme}. \citet{xtreme} introduce cross-lingual transfer gap to compare transferability. \citet{xlingual:mbert:iclr20} and \citet{elements:mbert} study the contribution of different components of mBERT to cross-lingual abilities. \citet{emerging:xlm:acl20} show that shared top layers of multilingual Transformers are necessary to achieve cross-lingual transfer. Similarly, \citet{muller2021first} understand mBERT as the stacking of a multilingual encoder followed by a language-agnostic predictor. \citet{lauscher-etal-2020-zero} study the limitations of zero-shot transfer. \citet{turc2021revisiting} find that some languages are more universally transferable than English. Multilingual Transformers are also analyzed in terms of language neutrality \citep{libovicky2019language,libovicky2020language}, word alignment \citep{simalign,xlmalign}, pre-training effects \citep{chai-etal-2022-cross,fujinuma-etal-2022-match}, chain-of-thought prompting \citep{brohan2023can,shi2023language}, etc. Different previous studies, our work focuses on measuring cross-lingual transferability of multilingual Transformers.

\section{Conclusion}

In this paper, we propose \our{}, a cross-lingual transferability metric for multilingual Transformers on sentence classification tasks. \our{} measures transferability by searching the minimum interlingual transfer gap while taking training errors into consideration. Our experiments show that \our{} better reflects cross-lingual transferability than baseline metrics. 
Through extensive experiments, we not only provide systematic comparisons of transferability but also have three findings on cross-lingual transfer. Besides, we present the transfer direction ranking task that evaluates transferability metrics without end-task data.

\paragraph{Limitations}
\our{} utilizes parallel sentences to measure cross-lingual transferability. The quality of parallel sentences or translated end-task examples affects the quality of resulting \our{} scores. Therefore, researchers should carefully select the parallel dataset to compute \our{} to ensure that \our{} measures cross-lingual transferability well. One of the potential risks is that \our{} can produce misleading cross-lingual transferability scores if poisoned translations are used.

\paragraph{Future work} For future work, we would like to employ \our{} to analyze multilingual Transformers on more NLP tasks such as language generation. Besides, how the quality of translated examples and parallel sentences affect our metric is also worth studying.

\bibliographystyle{plainnat}
\bibliography{xlme}

\begin{thebibliography}{41}
\providecommand{\natexlab}[1]{#1}
\providecommand{\url}[1]{\texttt{#1}}
\expandafter\ifx\csname urlstyle\endcsname\relax
  \providecommand{\doi}[1]{doi: #1}\else
  \providecommand{\doi}{doi: \begingroup \urlstyle{rm}\Url}\fi

\bibitem[Artetxe et~al.(2020)Artetxe, Ruder, and Yogatama]{xquad}
Mikel Artetxe, Sebastian Ruder, and Dani Yogatama.
\newblock On the cross-lingual transferability of monolingual representations.
\newblock In \emph{Proceedings of the 58th Annual Meeting of the Association
  for Computational Linguistics}, pages 4623--4637, Online, July 2020.
  Association for Computational Linguistics.
\newblock \doi{10.18653/v1/2020.acl-main.421}.
\newblock URL \url{https://www.aclweb.org/anthology/2020.acl-main.421}.

\bibitem[Brohan et~al.(2023)Brohan, Chebotar, Finn, Hausman, Herzog, Ho, Ibarz,
  Irpan, Jang, Julian, et~al.]{brohan2023can}
Anthony Brohan, Yevgen Chebotar, Chelsea Finn, Karol Hausman, Alexander Herzog,
  Daniel Ho, Julian Ibarz, Alex Irpan, Eric Jang, Ryan Julian, et~al.
\newblock Do as i can, not as i say: Grounding language in robotic affordances.
\newblock In \emph{Conference on Robot Learning}, pages 287--318. PMLR, 2023.

\bibitem[Chai et~al.(2022)Chai, Liang, and Duan]{chai-etal-2022-cross}
Yuan Chai, Yaobo Liang, and Nan Duan.
\newblock Cross-lingual ability of multilingual masked language models: A study
  of language structure.
\newblock In \emph{Proceedings of the 60th Annual Meeting of the Association
  for Computational Linguistics (Volume 1: Long Papers)}, pages 4702--4712,
  Dublin, Ireland, May 2022. Association for Computational Linguistics.
\newblock \doi{10.18653/v1/2022.acl-long.322}.
\newblock URL \url{https://aclanthology.org/2022.acl-long.322}.

\bibitem[Chi et~al.(2021{\natexlab{a}})Chi, Dong, Wei, Yang, Singhal, Wang,
  Song, Mao, Huang, and Zhou]{infoxlm}
Zewen Chi, Li~Dong, Furu Wei, Nan Yang, Saksham Singhal, Wenhui Wang, Xia Song,
  Xian-Ling Mao, Heyan Huang, and Ming Zhou.
\newblock {I}nfo{XLM}: An information-theoretic framework for cross-lingual
  language model pre-training.
\newblock In \emph{Proceedings of the 2021 Conference of the North American
  Chapter of the Association for Computational Linguistics: Human Language
  Technologies}, pages 3576--3588, Online, June 2021{\natexlab{a}}. Association
  for Computational Linguistics.
\newblock \doi{10.18653/v1/2021.naacl-main.280}.
\newblock URL \url{https://www.aclweb.org/anthology/2021.naacl-main.280}.

\bibitem[Chi et~al.(2021{\natexlab{b}})Chi, Dong, Zheng, Huang, Mao, Huang, and
  Wei]{xlmalign}
Zewen Chi, Li~Dong, Bo~Zheng, Shaohan Huang, Xian-Ling Mao, Heyan Huang, and
  Furu Wei.
\newblock Improving pretrained cross-lingual language models via self-labeled
  word alignment.
\newblock In \emph{Proceedings of the 59th Annual Meeting of the Association
  for Computational Linguistics and the 11th International Joint Conference on
  Natural Language Processing (Volume 1: Long Papers)}, pages 3418--3430,
  Online, August 2021{\natexlab{b}}. Association for Computational Linguistics.
\newblock \doi{10.18653/v1/2021.acl-long.265}.
\newblock URL \url{https://aclanthology.org/2021.acl-long.265}.

\bibitem[Chowdhery et~al.(2022)Chowdhery, Narang, Devlin, Bosma, Mishra,
  Roberts, Barham, Chung, Sutton, Gehrmann, et~al.]{palm}
Aakanksha Chowdhery, Sharan Narang, Jacob Devlin, Maarten Bosma, Gaurav Mishra,
  Adam Roberts, Paul Barham, Hyung~Won Chung, Charles Sutton, Sebastian
  Gehrmann, et~al.
\newblock Palm: Scaling language modeling with pathways.
\newblock \emph{arXiv preprint arXiv:2204.02311}, 2022.

\bibitem[Chung et~al.(2023)Chung, Garcia, Roberts, Tay, Firat, Narang, and
  Constant]{chung2023unimax}
Hyung~Won Chung, Xavier Garcia, Adam Roberts, Yi~Tay, Orhan Firat, Sharan
  Narang, and Noah Constant.
\newblock Unimax: Fairer and more effective language sampling for large-scale
  multilingual pretraining.
\newblock In \emph{The Eleventh International Conference on Learning
  Representations}, 2023.
\newblock URL \url{https://openreview.net/forum?id=kXwdL1cWOAi}.

\bibitem[Conneau and Lample(2019)]{xlm}
Alexis Conneau and Guillaume Lample.
\newblock Cross-lingual language model pretraining.
\newblock In \emph{Advances in Neural Information Processing Systems}, pages
  7057--7067. Curran Associates, Inc., 2019.
\newblock URL
  \url{http://papers.nips.cc/paper/8928-cross-lingual-language-model-pretraining.pdf}.

\bibitem[Conneau et~al.(2018)Conneau, Rinott, Lample, Williams, Bowman,
  Schwenk, and Stoyanov]{xnli}
Alexis Conneau, Ruty Rinott, Guillaume Lample, Adina Williams, Samuel Bowman,
  Holger Schwenk, and Veselin Stoyanov.
\newblock {XNLI}: Evaluating cross-lingual sentence representations.
\newblock In \emph{Proceedings of the 2018 Conference on Empirical Methods in
  Natural Language Processing}, pages 2475--2485, Brussels, Belgium,
  October-November 2018. Association for Computational Linguistics.
\newblock \doi{10.18653/v1/D18-1269}.
\newblock URL \url{https://www.aclweb.org/anthology/D18-1269}.

\bibitem[Conneau et~al.(2020{\natexlab{a}})Conneau, Khandelwal, Goyal,
  Chaudhary, Wenzek, Guzm{\'a}n, Grave, Ott, Zettlemoyer, and Stoyanov]{xlmr}
Alexis Conneau, Kartikay Khandelwal, Naman Goyal, Vishrav Chaudhary, Guillaume
  Wenzek, Francisco Guzm{\'a}n, Edouard Grave, Myle Ott, Luke Zettlemoyer, and
  Veselin Stoyanov.
\newblock Unsupervised cross-lingual representation learning at scale.
\newblock In \emph{Proceedings of the 58th Annual Meeting of the Association
  for Computational Linguistics}, pages 8440--8451, Online, July
  2020{\natexlab{a}}. Association for Computational Linguistics.
\newblock URL \url{https://www.aclweb.org/anthology/2020.acl-main.747}.

\bibitem[Conneau et~al.(2020{\natexlab{b}})Conneau, Wu, Li, Zettlemoyer, and
  Stoyanov]{emerging:xlm:acl20}
Alexis Conneau, Shijie Wu, Haoran Li, Luke Zettlemoyer, and Veselin Stoyanov.
\newblock Emerging cross-lingual structure in pretrained language models.
\newblock In \emph{Proceedings of the 58th Annual Meeting of the Association
  for Computational Linguistics}, pages 6022--6034, Online, July
  2020{\natexlab{b}}. Association for Computational Linguistics.
\newblock \doi{10.18653/v1/2020.acl-main.536}.
\newblock URL \url{https://www.aclweb.org/anthology/2020.acl-main.536}.

\bibitem[Devlin et~al.(2019)Devlin, Chang, Lee, and Toutanova]{bert}
Jacob Devlin, Ming-Wei Chang, Kenton Lee, and Kristina Toutanova.
\newblock {BERT}: Pre-training of deep bidirectional transformers for language
  understanding.
\newblock In \emph{Proceedings of the 2019 Conference of the North {A}merican
  Chapter of the Association for Computational Linguistics}, pages 4171--4186,
  Minneapolis, Minnesota, June 2019. Association for Computational Linguistics.
\newblock \doi{10.18653/v1/N19-1423}.
\newblock URL \url{https://www.aclweb.org/anthology/N19-1423}.

\bibitem[Dufter and Schutze(2020)]{elements:mbert}
Philipp Dufter and Hinrich Schutze.
\newblock Identifying necessary elements for {BERT}'s multilinguality.
\newblock \emph{ArXiv}, abs/2005.00396, 2020.

\bibitem[Fang et~al.(2021)Fang, Wang, Gan, Sun, and Liu]{filter}
Yuwei Fang, Shuohang Wang, Zhe Gan, Siqi Sun, and Jingjing Liu.
\newblock Filter: An enhanced fusion method for cross-lingual language
  understanding.
\newblock In \emph{Proceedings of the AAAI Conference on Artificial
  Intelligence}, volume~35, pages 12776--12784, 2021.

\bibitem[Feng et~al.(2022)Feng, Yang, Cer, Arivazhagan, and Wang]{labse}
Fangxiaoyu Feng, Yinfei Yang, Daniel Cer, Naveen Arivazhagan, and Wei Wang.
\newblock Language-agnostic {BERT} sentence embedding.
\newblock In \emph{Proceedings of the 60th Annual Meeting of the Association
  for Computational Linguistics (Volume 1: Long Papers)}, pages 878--891,
  Dublin, Ireland, May 2022. Association for Computational Linguistics.
\newblock \doi{10.18653/v1/2022.acl-long.62}.
\newblock URL \url{https://aclanthology.org/2022.acl-long.62}.

\bibitem[Fujinuma et~al.(2022)Fujinuma, Boyd-Graber, and
  Kann]{fujinuma-etal-2022-match}
Yoshinari Fujinuma, Jordan Boyd-Graber, and Katharina Kann.
\newblock Match the script, adapt if multilingual: Analyzing the effect of
  multilingual pretraining on cross-lingual transferability.
\newblock In \emph{Proceedings of the 60th Annual Meeting of the Association
  for Computational Linguistics (Volume 1: Long Papers)}, pages 1500--1512,
  Dublin, Ireland, May 2022. Association for Computational Linguistics.
\newblock \doi{10.18653/v1/2022.acl-long.106}.
\newblock URL \url{https://aclanthology.org/2022.acl-long.106}.

\bibitem[Goyal et~al.(2022)Goyal, Gao, Chaudhary, Chen, Wenzek, Ju, Krishnan,
  Ranzato, Guzmán, and Fan]{flores101}
Naman Goyal, Cynthia Gao, Vishrav Chaudhary, Peng-Jen Chen, Guillaume Wenzek,
  Da~Ju, Sanjana Krishnan, Marc’Aurelio Ranzato, Francisco Guzmán, and
  Angela Fan.
\newblock {The Flores-101 Evaluation Benchmark for Low-Resource and
  Multilingual Machine Translation}.
\newblock \emph{Transactions of the Association for Computational Linguistics},
  10:\penalty0 522--538, 05 2022.
\newblock ISSN 2307-387X.
\newblock \doi{10.1162/tacl_a_00474}.
\newblock URL \url{https://doi.org/10.1162/tacl\_a\_00474}.

\bibitem[Hu et~al.(2020{\natexlab{a}})Hu, Johnson, Firat, Siddhant, and
  Neubig]{hu2020explicit}
Junjie Hu, Melvin Johnson, Orhan Firat, Aditya Siddhant, and Graham Neubig.
\newblock Explicit alignment objectives for multilingual bidirectional
  encoders.
\newblock \emph{arXiv preprint arXiv:2010.07972}, 2020{\natexlab{a}}.

\bibitem[Hu et~al.(2020{\natexlab{b}})Hu, Ruder, Siddhant, Neubig, Firat, and
  Johnson]{xtreme}
Junjie Hu, Sebastian Ruder, Aditya Siddhant, Graham Neubig, Orhan Firat, and
  Melvin Johnson.
\newblock {XTREME}: A massively multilingual multi-task benchmark for
  evaluating cross-lingual generalization.
\newblock \emph{arXiv preprint arXiv:2003.11080}, 2020{\natexlab{b}}.

\bibitem[Jalili~Sabet et~al.(2020)Jalili~Sabet, Dufter, Yvon, and
  Sch{\"u}tze]{simalign}
Masoud Jalili~Sabet, Philipp Dufter, Fran{\c{c}}ois Yvon, and Hinrich
  Sch{\"u}tze.
\newblock {S}im{A}lign: High quality word alignments without parallel training
  data using static and contextualized embeddings.
\newblock In \emph{Findings of the Association for Computational Linguistics:
  EMNLP 2020}, pages 1627--1643, Online, November 2020. Association for
  Computational Linguistics.
\newblock \doi{10.18653/v1/2020.findings-emnlp.147}.
\newblock URL \url{https://www.aclweb.org/anthology/2020.findings-emnlp.147}.

\bibitem[K et~al.(2020)K, Wang, Mayhew, and Roth]{xlingual:mbert:iclr20}
Karthikeyan K, Zihan Wang, Stephen Mayhew, and Dan Roth.
\newblock Cross-lingual ability of multilingual bert: An empirical study.
\newblock In \emph{International Conference on Learning Representations}, 2020.
\newblock URL \url{https://openreview.net/forum?id=HJeT3yrtDr}.

\bibitem[Lauscher et~al.(2020)Lauscher, Ravishankar, Vuli{\'c}, and
  Glava{\v{s}}]{lauscher-etal-2020-zero}
Anne Lauscher, Vinit Ravishankar, Ivan Vuli{\'c}, and Goran Glava{\v{s}}.
\newblock From zero to hero: {O}n the limitations of zero-shot language
  transfer with multilingual {T}ransformers.
\newblock In \emph{Proceedings of the 2020 Conference on Empirical Methods in
  Natural Language Processing (EMNLP)}, pages 4483--4499, Online, November
  2020. Association for Computational Linguistics.
\newblock \doi{10.18653/v1/2020.emnlp-main.363}.
\newblock URL \url{https://aclanthology.org/2020.emnlp-main.363}.

\bibitem[Libovick{\`y} et~al.(2019)Libovick{\`y}, Rosa, and
  Fraser]{libovicky2019language}
Jind{\v{r}}ich Libovick{\`y}, Rudolf Rosa, and Alexander Fraser.
\newblock How language-neutral is multilingual bert?
\newblock \emph{arXiv preprint arXiv:1911.03310}, 2019.

\bibitem[Libovick{\`y} et~al.(2020)Libovick{\`y}, Rosa, and
  Fraser]{libovicky2020language}
Jind{\v{r}}ich Libovick{\`y}, Rudolf Rosa, and Alexander Fraser.
\newblock On the language neutrality of pre-trained multilingual
  representations.
\newblock In \emph{Findings of the Association for Computational Linguistics:
  EMNLP 2020}, pages 1663--1674, 2020.

\bibitem[Liu et~al.(2020)Liu, Gu, Goyal, Li, Edunov, Ghazvininejad, Lewis, and
  Zettlemoyer]{mbart}
Yinhan Liu, Jiatao Gu, Naman Goyal, Xian Li, Sergey Edunov, Marjan
  Ghazvininejad, Mike Lewis, and Luke Zettlemoyer.
\newblock Multilingual denoising pre-training for neural machine translation.
\newblock \emph{arXiv preprint arXiv:2001.08210}, 2020.

\bibitem[Luo et~al.(2020)Luo, Wang, Liu, Liu, Bi, Huang, Huang, and Si]{veco}
Fuli Luo, Wei Wang, Jiahao Liu, Yijia Liu, Bin Bi, Songfang Huang, Fei Huang,
  and Luo Si.
\newblock {VECO}: Variable encoder-decoder pre-training for cross-lingual
  understanding and generation.
\newblock \emph{arXiv preprint arXiv:2010.16046}, 2020.

\bibitem[Muller et~al.(2021)Muller, Elazar, Sagot, and Seddah]{muller2021first}
Benjamin Muller, Yanai Elazar, Beno{\^\i}t Sagot, and Djam{\'e} Seddah.
\newblock First align, then predict: Understanding the cross-lingual ability of
  multilingual bert.
\newblock In \emph{Proceedings of the 16th Conference of the European Chapter
  of the Association for Computational Linguistics: Main Volume}, pages
  2214--2231, 2021.

\bibitem[Ouyang et~al.(2020)Ouyang, Wang, Pang, Sun, Tian, Wu, and
  Wang]{erniem}
Xuan Ouyang, Shuohuan Wang, Chao Pang, Yu~Sun, Hao Tian, Hua Wu, and Haifeng
  Wang.
\newblock Ernie-m: Enhanced multilingual representation by aligning
  cross-lingual semantics with monolingual corpora.
\newblock \emph{arXiv preprint arXiv:2012.15674}, 2020.

\bibitem[Pires et~al.(2019)Pires, Schlinger, and
  Garrette]{pires2019multilingual}
Telmo Pires, Eva Schlinger, and Dan Garrette.
\newblock How multilingual is multilingual {BERT}?
\newblock In \emph{Proceedings of the 57th Annual Meeting of the Association
  for Computational Linguistics}, pages 4996--5001, Florence, Italy, July 2019.
  Association for Computational Linguistics.
\newblock \doi{10.18653/v1/P19-1493}.
\newblock URL \url{https://www.aclweb.org/anthology/P19-1493}.

\bibitem[Scao et~al.(2022)Scao, Fan, Akiki, Pavlick, Ili{\'c}, Hesslow,
  Castagn{\'e}, Luccioni, Yvon, Gall{\'e}, et~al.]{scao2022bloom}
Teven~Le Scao, Angela Fan, Christopher Akiki, Ellie Pavlick, Suzana Ili{\'c},
  Daniel Hesslow, Roman Castagn{\'e}, Alexandra~Sasha Luccioni, Fran{\c{c}}ois
  Yvon, Matthias Gall{\'e}, et~al.
\newblock Bloom: A 176b-parameter open-access multilingual language model.
\newblock \emph{arXiv preprint arXiv:2211.05100}, 2022.

\bibitem[Shi et~al.(2023)Shi, Suzgun, Freitag, Wang, Srivats, Vosoughi, Chung,
  Tay, Ruder, Zhou, Das, and Wei]{shi2023language}
Freda Shi, Mirac Suzgun, Markus Freitag, Xuezhi Wang, Suraj Srivats, Soroush
  Vosoughi, Hyung~Won Chung, Yi~Tay, Sebastian Ruder, Denny Zhou, Dipanjan Das,
  and Jason Wei.
\newblock Language models are multilingual chain-of-thought reasoners.
\newblock In \emph{The Eleventh International Conference on Learning
  Representations}, 2023.
\newblock URL \url{https://openreview.net/forum?id=fR3wGCk-IXp}.

\bibitem[Turc et~al.(2021)Turc, Lee, Eisenstein, Chang, and
  Toutanova]{turc2021revisiting}
Iulia Turc, Kenton Lee, Jacob Eisenstein, Ming-Wei Chang, and Kristina
  Toutanova.
\newblock Revisiting the primacy of english in zero-shot cross-lingual
  transfer.
\newblock \emph{arXiv preprint arXiv:2106.16171}, 2021.

\bibitem[Vaswani et~al.(2017)Vaswani, Shazeer, Parmar, Uszkoreit, Jones, Gomez,
  Kaiser, and Polosukhin]{transformer}
Ashish Vaswani, Noam Shazeer, Niki Parmar, Jakob Uszkoreit, Llion Jones,
  Aidan~N Gomez, {\L}ukasz Kaiser, and Illia Polosukhin.
\newblock Attention is all you need.
\newblock In \emph{Advances in Neural Information Processing Systems}, pages
  5998--6008. Curran Associates, Inc., 2017.
\newblock URL
  \url{http://papers.nips.cc/paper/7181-attention-is-all-you-need.pdf}.

\bibitem[Wei et~al.(2021)Wei, Weng, Hu, Xing, Yu, and Luo]{hictl}
Xiangpeng Wei, Rongxiang Weng, Yue Hu, Luxi Xing, Heng Yu, and Weihua Luo.
\newblock On learning universal representations across languages.
\newblock In \emph{International Conference on Learning Representations}, 2021.
\newblock URL \url{https://openreview.net/forum?id=Uu1Nw-eeTxJ}.

\bibitem[Williams et~al.(2018)Williams, Nangia, and Bowman]{mnli2017}
Adina Williams, Nikita Nangia, and Samuel Bowman.
\newblock A broad-coverage challenge corpus for sentence understanding through
  inference.
\newblock In \emph{NAACL}, pages 1112--1122, New Orleans, Louisiana, June 2018.
\newblock \doi{10.18653/v1/N18-1101}.
\newblock URL \url{https://www.aclweb.org/anthology/N18-1101}.

\bibitem[Wu and Dredze(2019)]{wu2019beto}
Shijie Wu and Mark Dredze.
\newblock Beto, bentz, becas: The surprising cross-lingual effectiveness of
  {BERT}.
\newblock In \emph{Proceedings of the 2019 Conference on Empirical Methods in
  Natural Language Processing and the 9th International Joint Conference on
  Natural Language Processing}, pages 833--844, Hong Kong, China, November
  2019. Association for Computational Linguistics.
\newblock \doi{10.18653/v1/D19-1077}.
\newblock URL \url{https://www.aclweb.org/anthology/D19-1077}.

\bibitem[Xue et~al.(2021)Xue, Constant, Roberts, Kale, Al-Rfou, Siddhant,
  Barua, and Raffel]{mt5}
Linting Xue, Noah Constant, Adam Roberts, Mihir Kale, Rami Al-Rfou, Aditya
  Siddhant, Aditya Barua, and Colin Raffel.
\newblock m{T}5: A massively multilingual pre-trained text-to-text transformer.
\newblock In \emph{Proceedings of the 2021 Conference of the North American
  Chapter of the Association for Computational Linguistics: Human Language
  Technologies}, pages 483--498, Online, June 2021. Association for
  Computational Linguistics.
\newblock \doi{10.18653/v1/2021.naacl-main.41}.
\newblock URL \url{https://aclanthology.org/2021.naacl-main.41}.

\bibitem[Yang et~al.(2022)Yang, Chen, Zhou, and Li]{yang2022enhancing}
Huiyun Yang, Huadong Chen, Hao Zhou, and Lei Li.
\newblock Enhancing cross-lingual transfer by manifold mixup.
\newblock In \emph{International Conference on Learning Representations}, 2022.
\newblock URL \url{https://openreview.net/forum?id=OjPmfr9GkVv}.

\bibitem[Yang et~al.(2019)Yang, Zhang, Tar, and Baldridge]{pawsx}
Yinfei Yang, Yuan Zhang, Chris Tar, and Jason Baldridge.
\newblock {PAWS}-{X}: A cross-lingual adversarial dataset for paraphrase
  identification.
\newblock In \emph{Proceedings of the 2019 Conference on Empirical Methods in
  Natural Language Processing and the 9th International Joint Conference on
  Natural Language Processing (EMNLP-IJCNLP)}, pages 3687--3692, Hong Kong,
  China, November 2019. Association for Computational Linguistics.
\newblock \doi{10.18653/v1/D19-1382}.
\newblock URL \url{https://www.aclweb.org/anthology/D19-1382}.

\bibitem[Zhang et~al.(2017)Zhang, Bengio, Hardt, Recht, and
  Vinyals]{zhang2017understanding}
Chiyuan Zhang, Samy Bengio, Moritz Hardt, Benjamin Recht, and Oriol Vinyals.
\newblock Understanding deep learning requires rethinking generalization.
\newblock In \emph{International Conference on Learning Representations}, 2017.
\newblock URL \url{https://openreview.net/forum?id=Sy8gdB9xx}.

\bibitem[Zheng et~al.(2021)Zheng, Dong, Huang, Wang, Chi, Singhal, Che, Liu,
  Song, and Wei]{xtune}
Bo~Zheng, Li~Dong, Shaohan Huang, Wenhui Wang, Zewen Chi, Saksham Singhal,
  Wanxiang Che, Ting Liu, Xia Song, and Furu Wei.
\newblock Consistency regularization for cross-lingual fine-tuning.
\newblock In \emph{Proceedings of the 59th Annual Meeting of the Association
  for Computational Linguistics and the 11th International Joint Conference on
  Natural Language Processing (Volume 1: Long Papers)}, pages 3403--3417,
  Online, August 2021. Association for Computational Linguistics.
\newblock \doi{10.18653/v1/2021.acl-long.264}.
\newblock URL \url{https://aclanthology.org/2021.acl-long.264}.

\end{thebibliography}

\clearpage
\appendix

\section{Experiment details}

In our experiments, the data of XNLI and PAWS-X are from the XTREME \citep{xtreme} benchmark. The repository\footnote{\url{github.com/google-research/xtreme}} provides data, data processing scripts, and the license. For transfer direction ranking, we use the parallel data from FLORES-101~\citep{flores101}, and the repository\footnote{\url{github.com/facebookresearch/flores}} provides data and license. The multilingual Transformers are from Hugging Face\footnote{\url{huggingface.co}}. We implement the fine-tuning algorithms with PyTorch\footnote{\url{pytorch.org}}, and plot figures with matplotlib\footnote{\url{matplotlib.org}}. We fine-tune the multilingual Transformers with three random seeds on NVIDIA GeForce RTX 3090 GPUs. The hyperparameters of fine-tuning are shown in Table~\ref{table:fthparam}.

\begin{table}[h]
\caption{Hyperparameters for fine-tuning on XNLI, PAWS-X, and FLORES-101.}
\centering
\scalebox{0.9}{
\begin{tabular}{lrrr}
\toprule
\bf Hyperparameters & XNLI & PAWS-X & FLORES-101 \\ \midrule
\multicolumn{3}{l}{\textit{Common hyperparameters}} \\
Batch size & 32 & 32 & 32 \\
Learning rate & 5e-6,\{1,2\}e-5 & 5e-6,\{1,2\}e-5 & 2e-5 \\
LR schedule & Fixed & Fixed & Fixed\\
Warmup & 10\% & 10\% & 10\% \\
Weight decay & 0 & 0 & 0\\
Epochs & 40 & 60 & 32 \\ \midrule
\multicolumn{3}{l}{\textit{Hyperparameters for \textsc{Gaussian} and \textsc{xTune}}} \\
Gaussian mean & 0 & 0 & - \\
Gaussian std & 0.075 & 0.075 & - \\
KL Weight $\lambda_1$ & 1.0 & 1.0 & - \\
\bottomrule
\end{tabular}
}
\label{table:fthparam}
\end{table}

\section{Supplementary results}
\label{sec:app-results}

\paragraph{Transfer direction ranking}

Previous studies typically utilize MNLI~\citep{mnli2017} as the training data for XNLI, so we further perform transfer direction ranking where we fine-tune XLM-R on MNLI. The TDR accuracy is computed in $14$ transfer directions from English to the other languages, because MNLI only provides training data in English. The results are shown in Table~\ref{table:app-tdr}. \our{} achieves the best TDR accuracy.

\begin{table}[h]
\caption{Transfer direction ranking on XNLI where we fine-tune XLM-R on MNLI.}
\centering
\scalebox{0.9}{
\begin{tabular}{lcccc}
\toprule
\bf Metric & $L_2$ & \textsc{Dot} & \textsc{Cos} & \our{} \\ \midrule
\bf acc & 79.1 & 72.5 & 81.3 & \bf 87.9 \\
\bottomrule
\end{tabular}
}
\label{table:app-tdr}
\end{table}

\paragraph{Compare multilingual Transformers}
Table~\ref{table:app-cm1} and Table~\ref{table:app-cm2} provide the detailed \our{} scores of multilingual Transformers.

\begin{table*}[h]
\caption{Detailed \our{} scores of multilingual Transformers on XNLI transferring from English to other languages.}
\centering
\scalebox{0.85}{
\renewcommand\tabcolsep{4.2pt}
\begin{tabular}{l|ccccccccccccccc|c}
\toprule
\bf Model &  ar  &   bg  &   de  &   el   &   es  &   fr  &   hi  &   ru  &   sw  &   th  &   tr  &   ur  &   vi  &   zh  &  avg & test acc \\ \midrule
mBERT & 33.6 &  29.6 &  25.4 &  30.8 &  18.8 &  22.1 &  37.7 &  30.7 &  49.5 &  44.6 &  33.9 &  41.2 &  27.4 &  28.2 &  34.9 & 59.5 \\
XLM & 24.8 &  21.5 &  17.8 &  19.6 &  14.5 &  17.8 &  33.6 &  23.6 &  28.4 &  47.3 &  25.6 &  33.8 &  22.3 &  28.2 &  27.6 & 62.9 \\
XLM-R & 23.1 &  15.9 &  15.6 &  18.1 &  11.6 &  15.9 &  26.0 &  20.7 &  31.9 &  23.1 &  23.7 &  30.1 &  18.6 &  21.1 &  21.1 & 65.7 \\
InfoXLM & 21.0 &  15.4 &  13.9 &  15.2 &  10.7 &  13.7 &  22.7 &  18.9 &  27.6 &  19.4 &  19.9 &  26.4 &  17.2 &  18.2 &  20.0 & 66.6 \\
XLM-R$_\text{large}$ & 13.8 &   8.2 &   6.7 &   8.6 &   5.4 &   6.9 &  16.7 &  12.6 &  20.6 &  14.4 &  13.9 &  19.7 &  10.8 &  12.4 &  13.1 & 74.4 \\
\bottomrule
\end{tabular}
}
\label{table:app-cm1}
\end{table*}

\begin{table*}
\caption{Detailed \our{} scores of multilingual Transformers on PAWS-X transferring from English to other languages.}
\centering
\scalebox{0.85}{
\begin{tabular}{l|ccccccc|c}
\toprule
\bf Model &  de  &   es  &   fr  &   ja  &   ko  &   zh  &  avg & test acc \\ \midrule
mBERT & 14.7 &   9.8 &   8.1 &  26.3 &  25.8 &  23.1 &  21.6 & 74.1 \\
XLM & 10.9 &   6.3 &   5.6 &  33.1 &  38.1 &  22.6 &  23.3 & 75.4 \\
XLM-R & 10.5 &   6.8 &   6.6 &  21.3 &  20.3 &  18.4 &  16.8 & 80.1 \\
InfoXLM & 9.3 &   4.5 &   4.8 &  17.6 &  17.7 &  16.1 &  14.0 & 82.9 \\
XLM-R$_\text{large}$ & 8.5 &   5.3 &   5.4 &  15.1 &  15.9 &  14.5 &  12.9 & 85.1 \\
\bottomrule
\end{tabular}
}
\label{table:app-cm2}
\end{table*}

\paragraph{Compare fine-tuning algorithms} Table \ref{table:app-ca} presents the \our{} scores of various fine-tuning algorithms, where we fine-tune XLM-R on PAWS-X.

\begin{table*}
\caption{\our{} scores of various fine-tuning algorithms transferring from English to other languages, where we fine-tune XLM-R on PAWS-X.}
\centering
\scalebox{0.85}{
\begin{tabular}{l|ccccccc|c}
\toprule
\bf Algorithm &  de  &   es  &   fr  &   ja  &   ko  &   zh  &  avg & test acc \\ \midrule
\textsc{Vanilla} & 10.5 &   6.8 &   6.6 &  21.3 &  20.3 &  18.4 &  16.8 & 80.1 \\
\textsc{Gaussian} & 10.0 &   7.1 &   6.3 &  19.9 &  18.8 &  17.8 &  15.9 & 80.9 \\
\textsc{xTune} & 10.1 &   6.2 &   6.2 &  20.1 &  18.9 &  17.2 &  15.7 & 81.4 \\
\bottomrule
\end{tabular}
}
\label{table:app-ca}
\end{table*}

\begin{figure*}[t]
 \centering
 \includegraphics[width=0.7\textwidth]{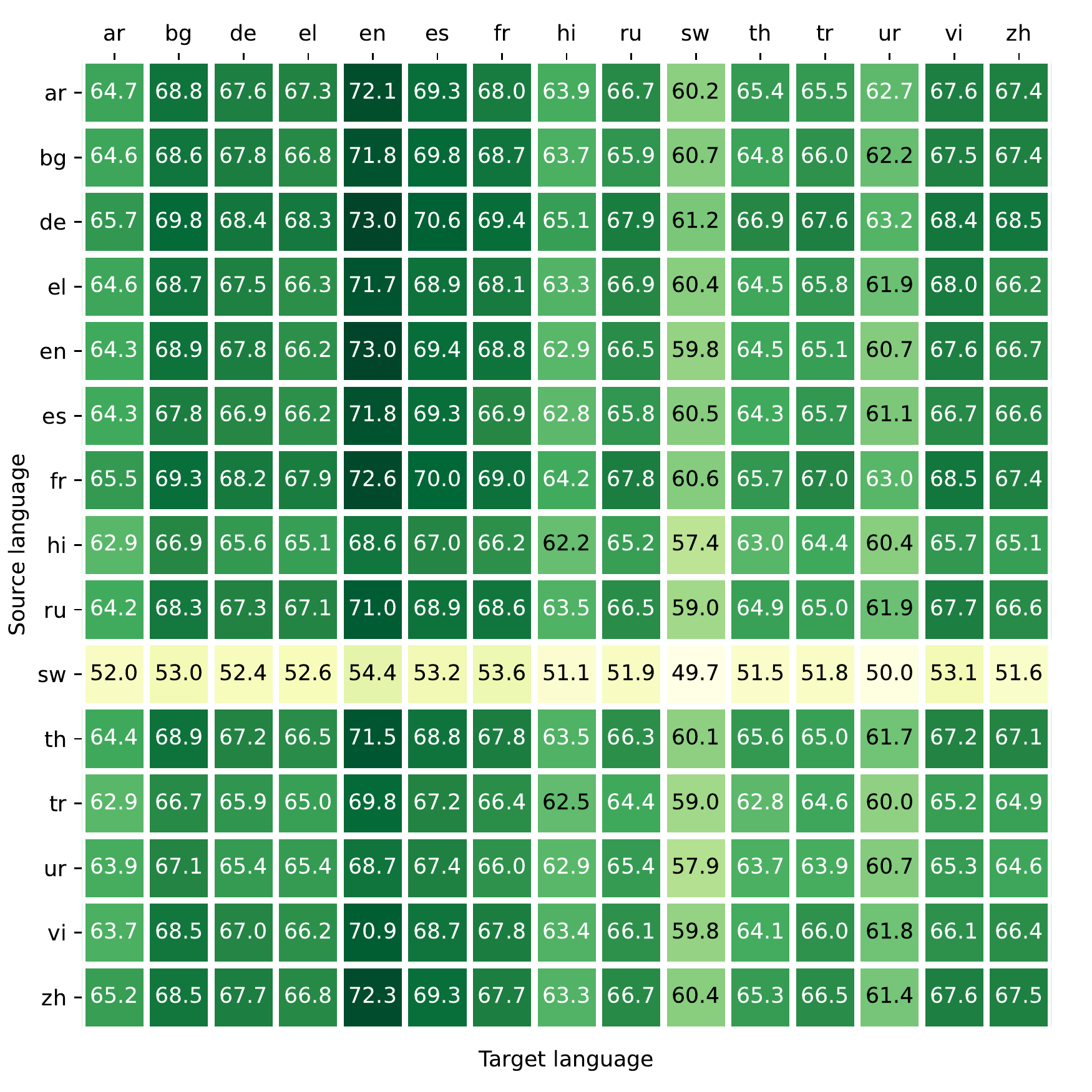}
\caption{Test accuracy of XLM-R in $15 \times 15$ transfer directions on XNLI.}
\label{fig:app-td1}
\end{figure*}

\begin{figure*}[t]
 \centering
 \includegraphics[width=0.7\textwidth]{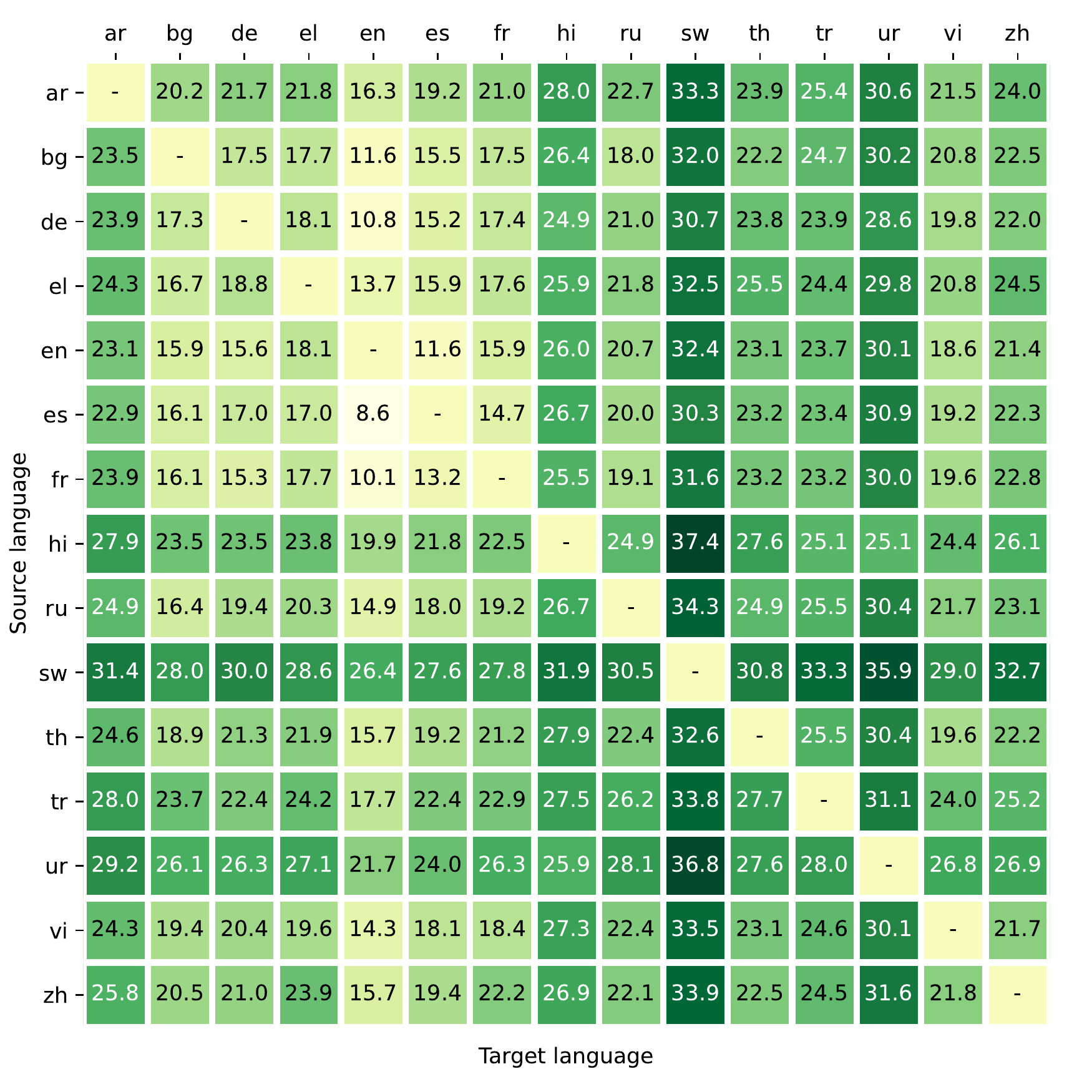}
\caption{\our{} of XLM-R in $15 \times 15$ transfer directions on XNLI.}
\label{fig:app-td2}
\end{figure*}

\begin{figure*}[t]
 \centering
 \includegraphics[width=0.7\textwidth]{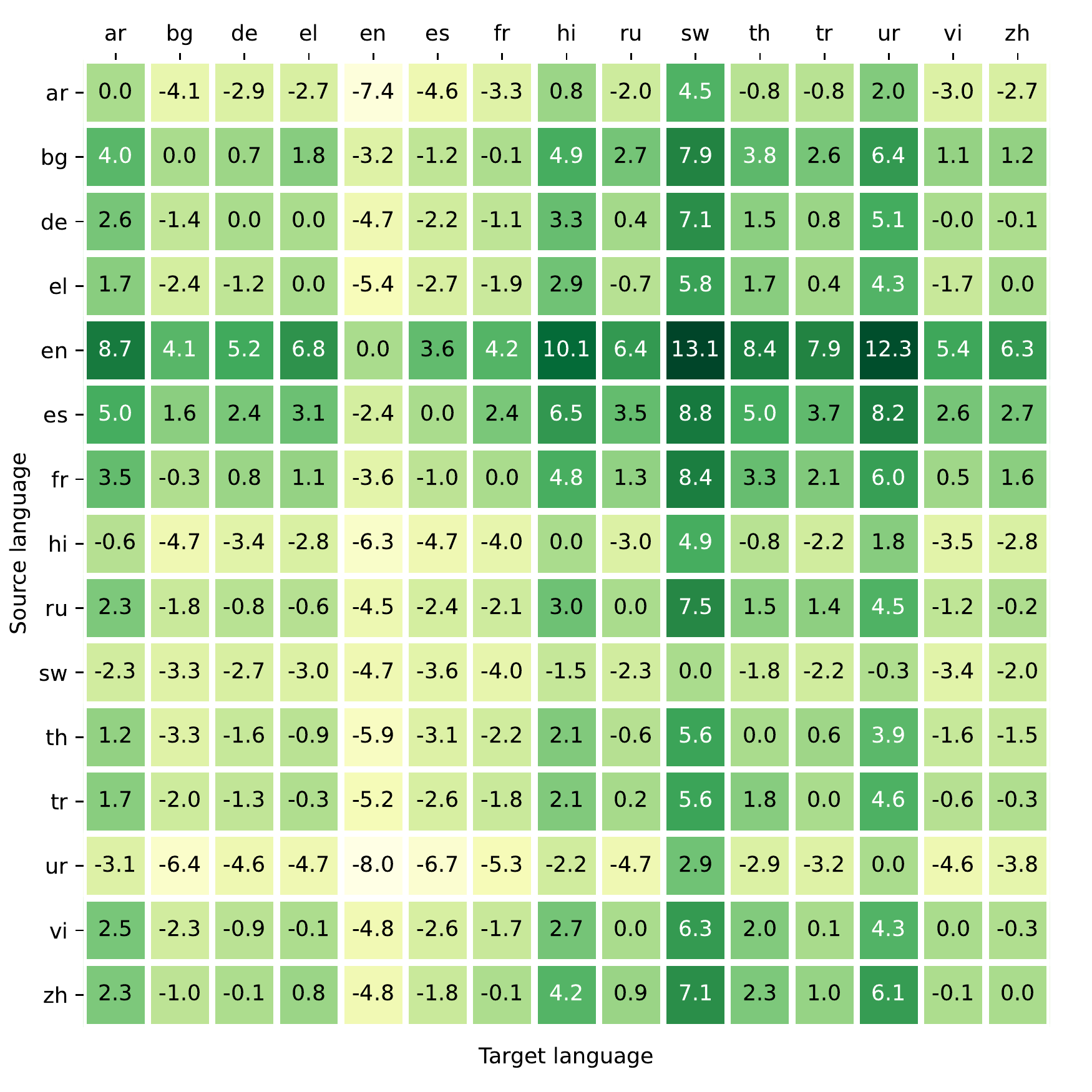}
\caption{Cross-lingual transfer gap of XLM-R in $15 \times 15$ transfer directions on XNLI.}
\label{fig:app-td3}
\end{figure*}

\paragraph{Compare transfer directions}
Figure \ref{fig:app-td1}, Figure \ref{fig:app-td2}, and Figure \ref{fig:app-td3} illustrate the test accuracy, \our{}, and cross-lingual transfer gap scores of XLM-R on XNLI in $15 \times 15$ transfer directions.

\paragraph{\our{} for larger-scale fine-tuning} To obtain XNLI models, previous methods \citep{xlmr,infoxlm} typically employ MNLI~\citep{mnli2017} as the training set for cross-lingual transfer, because MNLI provides much more data for training. We compute the \our{} of multilingual Transformers using MNLI. Since MNLI does not provide human-translated examples in other languages, we use both the MNLI data and XNLI validation data as the training data for fine-tuning. We fine-tune the models for $5$ epochs with a learning rate of $2 \times 10^{-5}$. Then, we compute expected \our{} on XNLI validation data because the data also serve as training data during fine-tuning. We use $\mathcal{E}' = 0.03$ and $\epsilon = 0.025$ when computing \our{}.
The \our{} scores are shown in Table~\ref{table:app-cm3}. Interestingly, comparing the results with Table~\ref{table:app-cm1}, it shows that the multilingual Transformers have very similar average \our{} scores with the scores in Table~\ref{table:app-cm1}, even though they achieve much better test accuracy because of larger-scale fine-tuning. It demonstrates that \our{} also works for larger-scale fine-tuning situations.

\begin{table*}[h]
\caption{\our{} scores of multilingual Transformers on XNLI transferring from English to other languages. The last column shows the test accuracy averaged with $14$ target languages.}
\centering
\scalebox{0.83}{
\renewcommand\tabcolsep{4.0pt}
\begin{tabular}{l|ccccccccccccccc|c}
\toprule
\bf Model &  ar  &   bg  &   de  &   el   &   es  &   fr  &   hi  &   ru  &   sw  &   th  &   tr  &   ur  &   vi  &   zh  &  avg & test acc \\ \midrule
mBERT & 31.0 &  28.3 &  23.7 &  29.9 &  18.9 &  20.8 &  35.9 &  27.5 &  46.2 &  44.5 &  35.1 &  38.3 &  25.4 &  25.5 &  33.1 & 63.3 \\
XLM & 23.5 &  20.4 &  19.0 &  19.9 &  15.5 &  17.1 &  33.5 &  22.7 &  28.7 &  43.7 &  27.3 &  33.7 &  23.2 &  27.4 &  27.4 & 67.6 \\
XLM-R &  23.2 &  17.3 &  16.6 &  17.8 &  13.5 &  13.9 &  25.7 &  18.8 &  34.1 &  22.7 &  22.0 &  29.3 &  18.8 &  20.9 &  22.7 & 71.7 \\
InfoXLM & 19.9 &  15.2 &  13.9 &  15.1 &  11.7 &  12.4 &  23.5 &  16.9 &  30.2 &  20.3 &  20.6 &  27.9 &  16.9 &  18.0 &  20.2 & 73.3 \\
XLM-R$_\text{large}$ & 17.1 &  10.4 &  10.8 &  13.2 &   8.0 &  10.7 &  19.5 &  13.7 &  29.7 &  17.3 &  18.0 &  23.5 &  14.7 &  15.3 &  17.1 & 75.2\\
\bottomrule
\end{tabular}
}
\label{table:app-cm3}
\end{table*}

\begin{figure*}[t!]
\centering
\begin{subfigure}[b]{1.0\textwidth}
    \begin{subfigure}[b]{0.32\textwidth}
     \centering
     \includegraphics[width=0.95\textwidth]{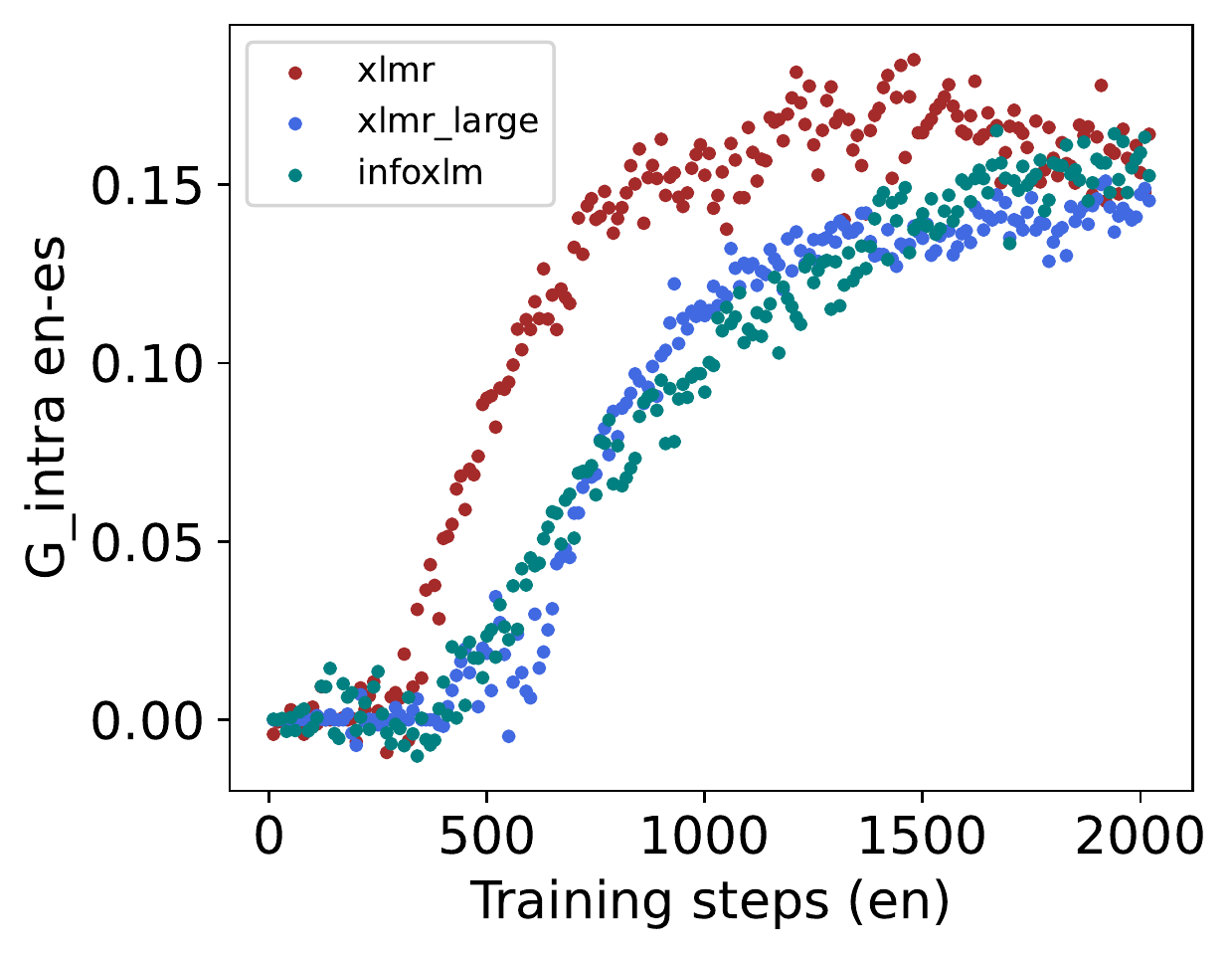}
    \end{subfigure}
    \hfill
    \begin{subfigure}[b]{0.32\textwidth}
     \centering
     \includegraphics[width=0.95\textwidth]{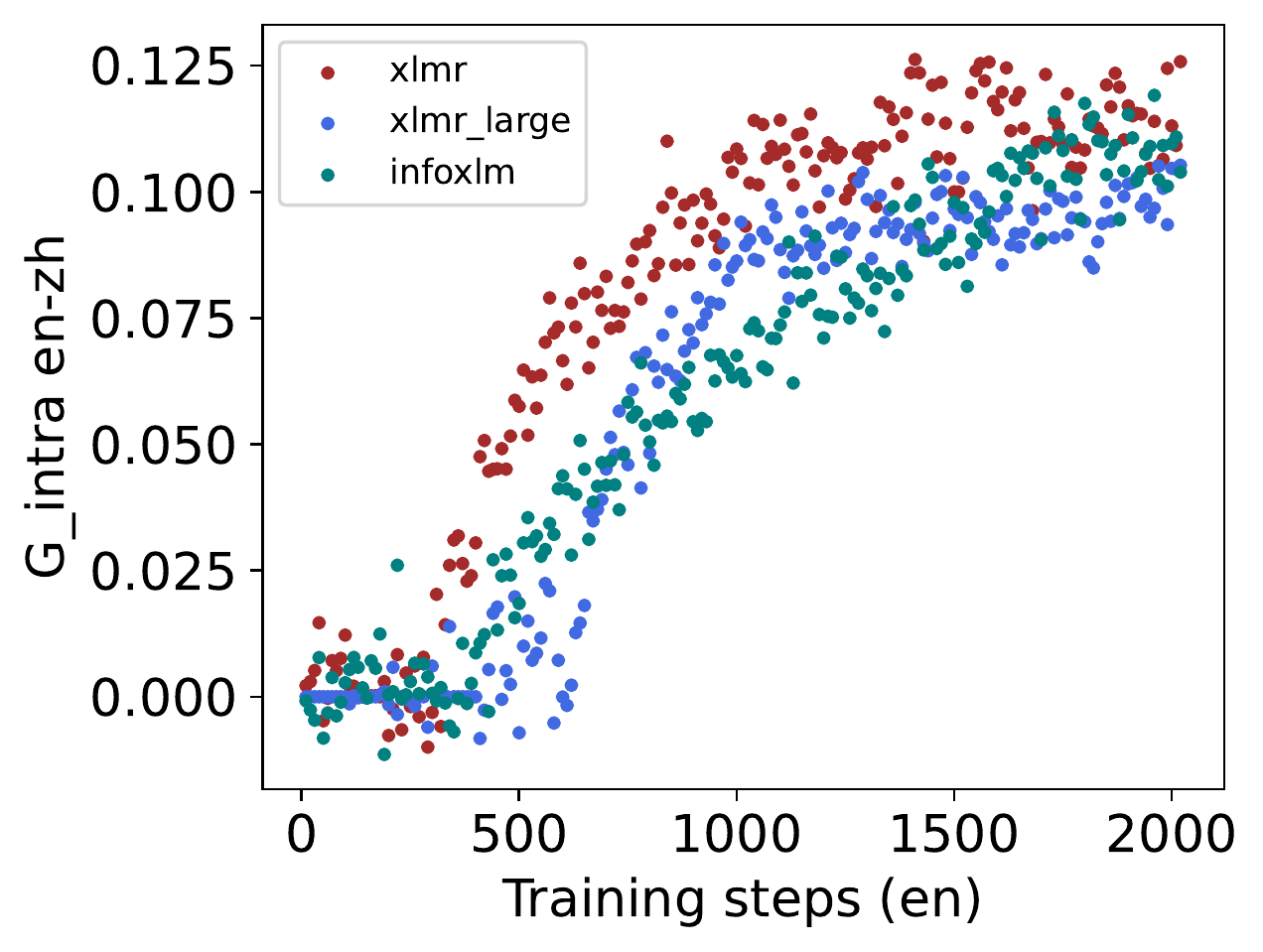}
    \end{subfigure}
    \hfill
    \begin{subfigure}[b]{0.32\textwidth}
     \centering
     \includegraphics[width=0.95\textwidth]{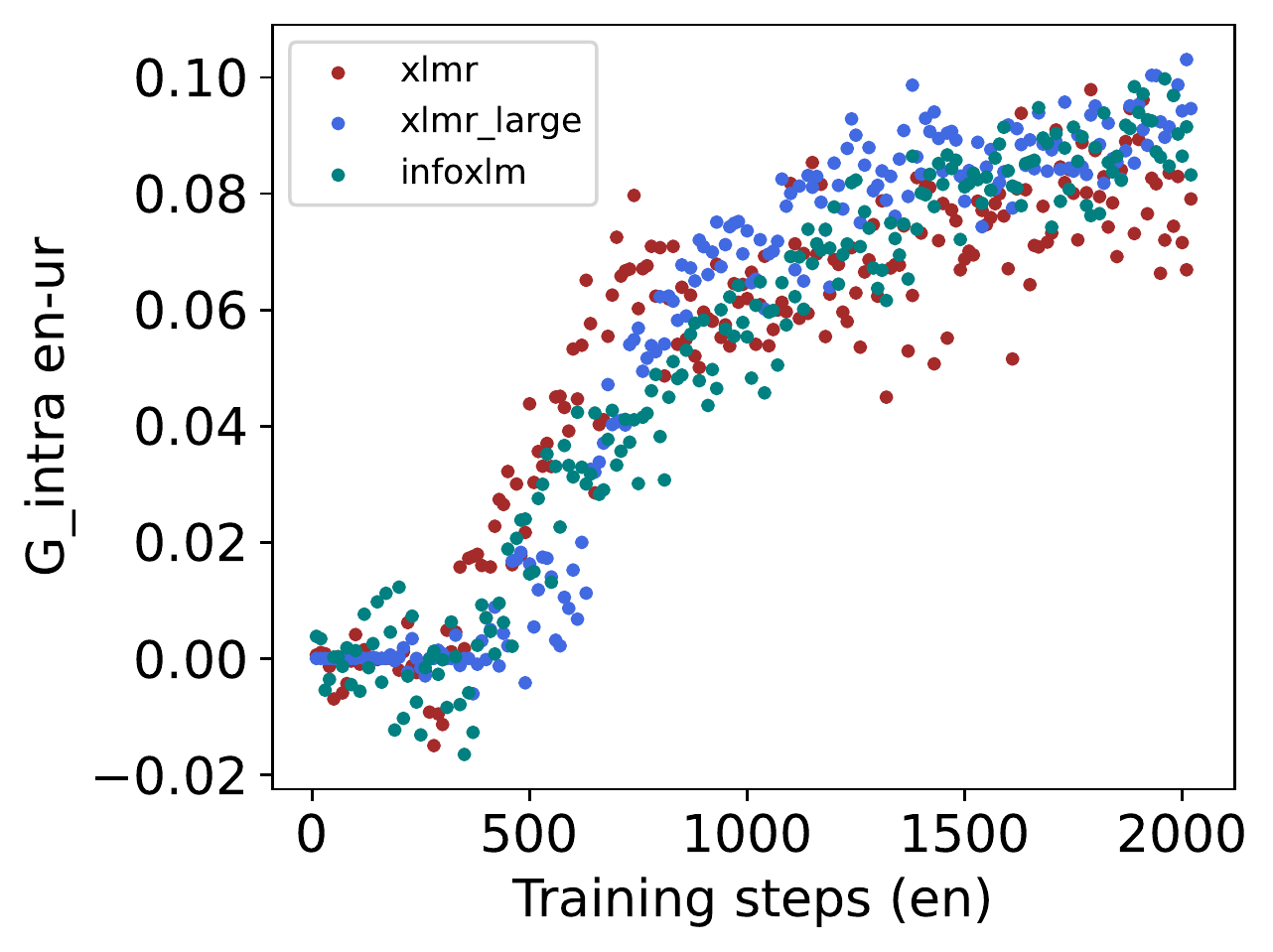}
    \end{subfigure}
\end{subfigure}
\begin{subfigure}[b]{1.0\textwidth}
    \begin{subfigure}[b]{0.32\textwidth}
     \centering
     \includegraphics[width=0.98\textwidth]{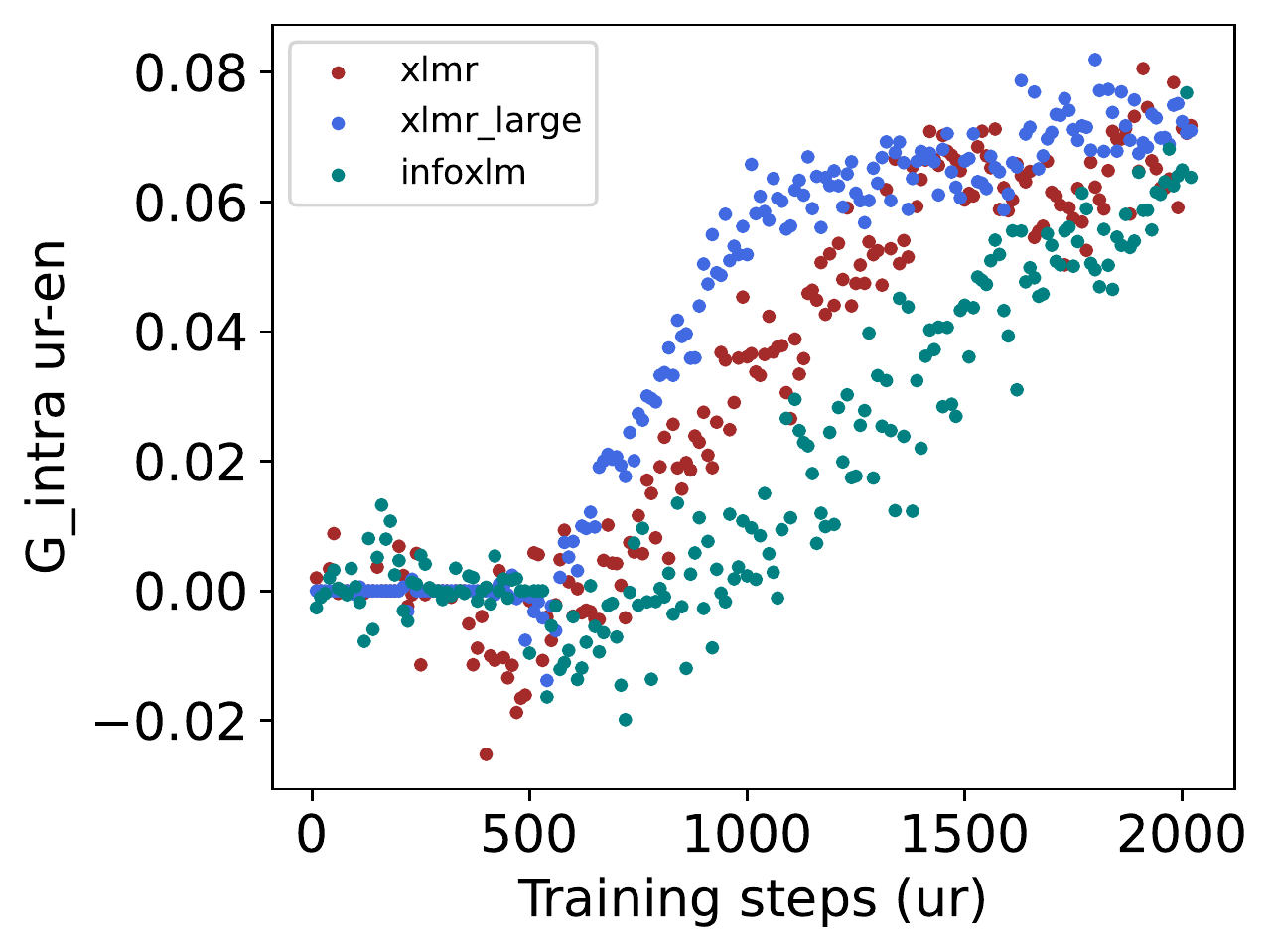}
    \end{subfigure}
    \hfill
    \begin{subfigure}[b]{0.32\textwidth}
     \centering
     \includegraphics[width=0.97\textwidth]{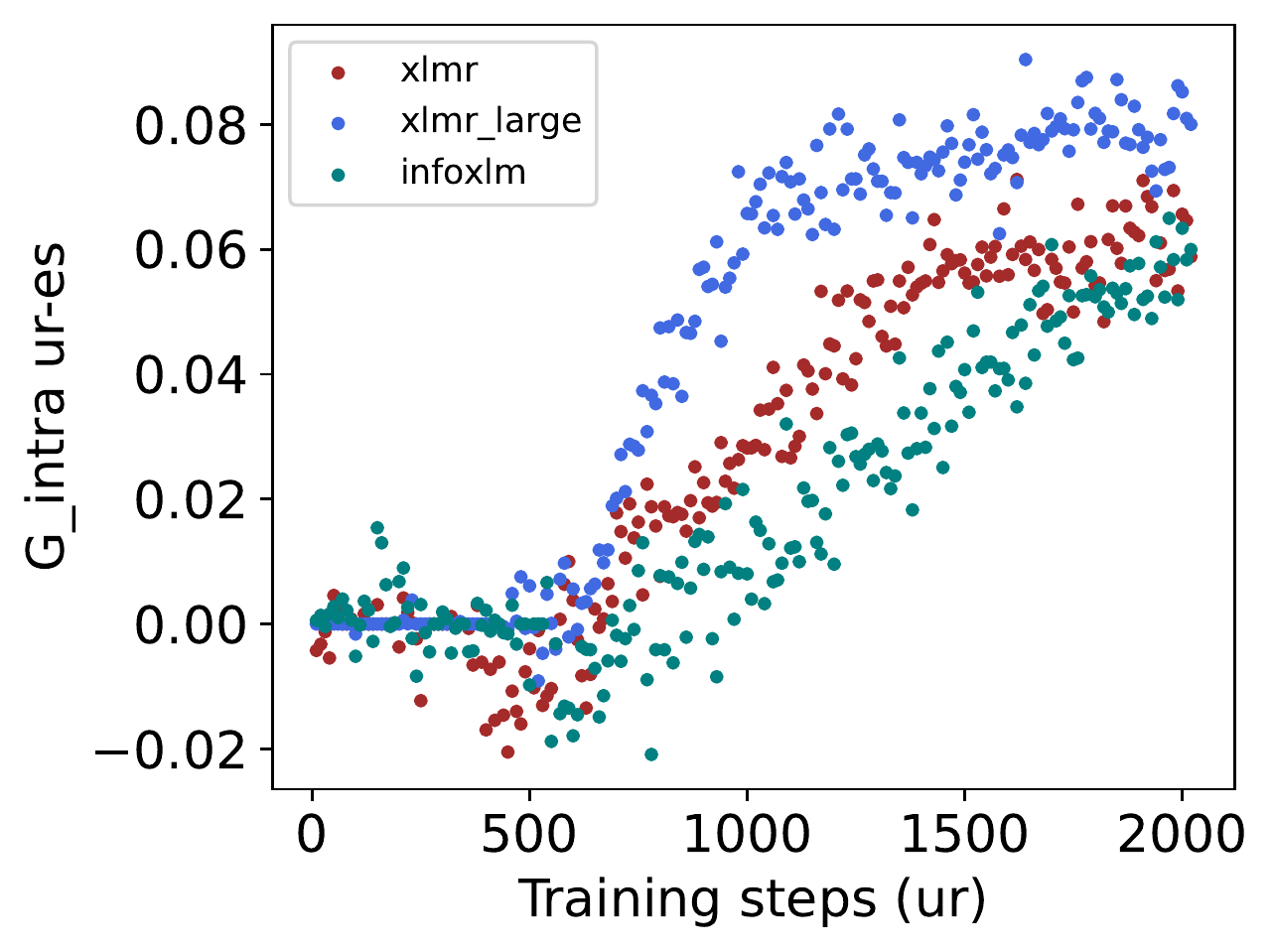}
    \end{subfigure}
    \hfill
    \begin{subfigure}[b]{0.32\textwidth}
     \centering
     \includegraphics[width=0.96\textwidth]{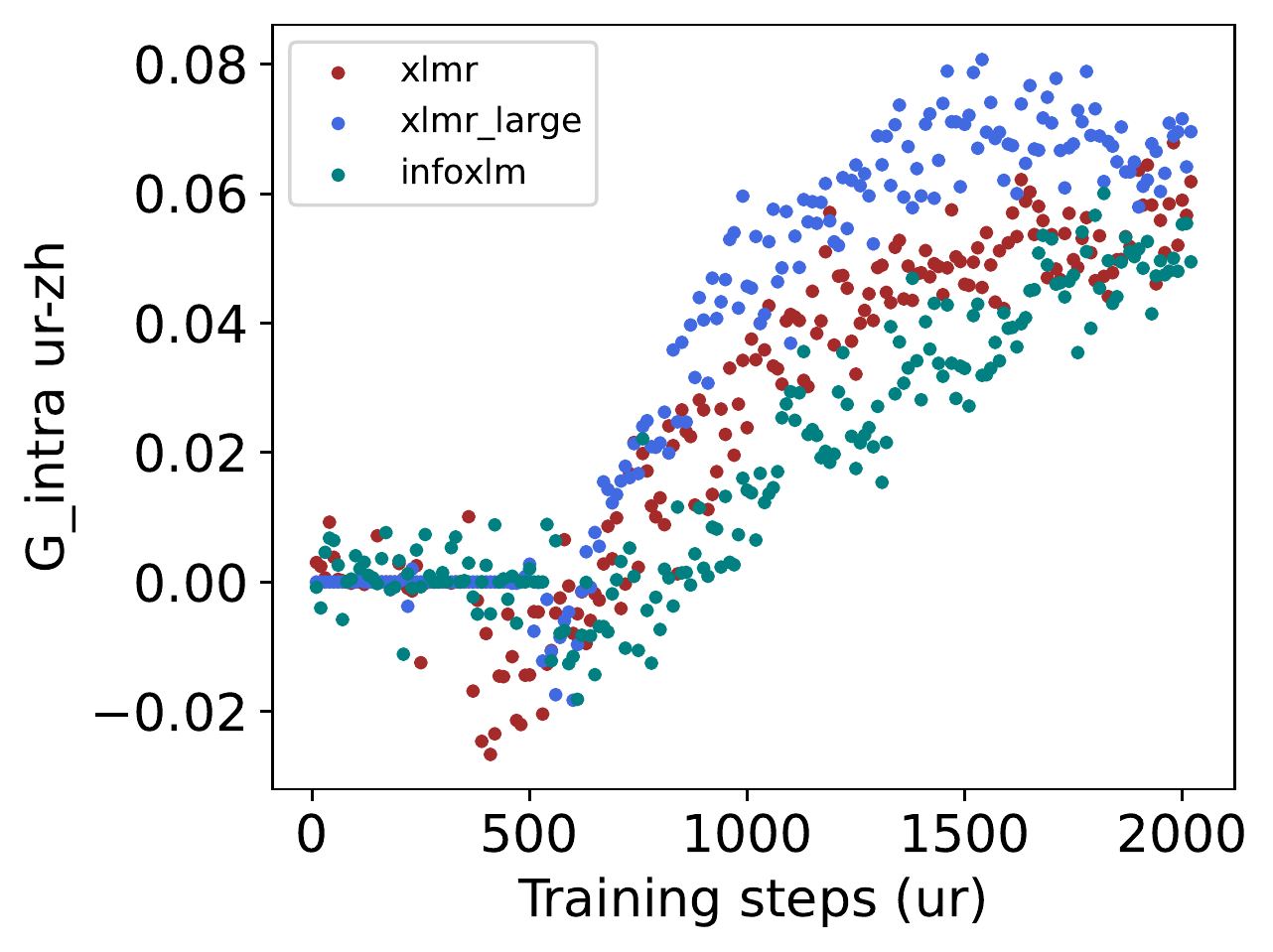}
    \end{subfigure}
    \caption{Intralingual generalization gap scores of various multilingual Transformers.}
\end{subfigure}
\begin{subfigure}[b]{1.0\textwidth}
    \begin{subfigure}[b]{0.33\textwidth}
     \centering
     \includegraphics[width=0.95\textwidth]{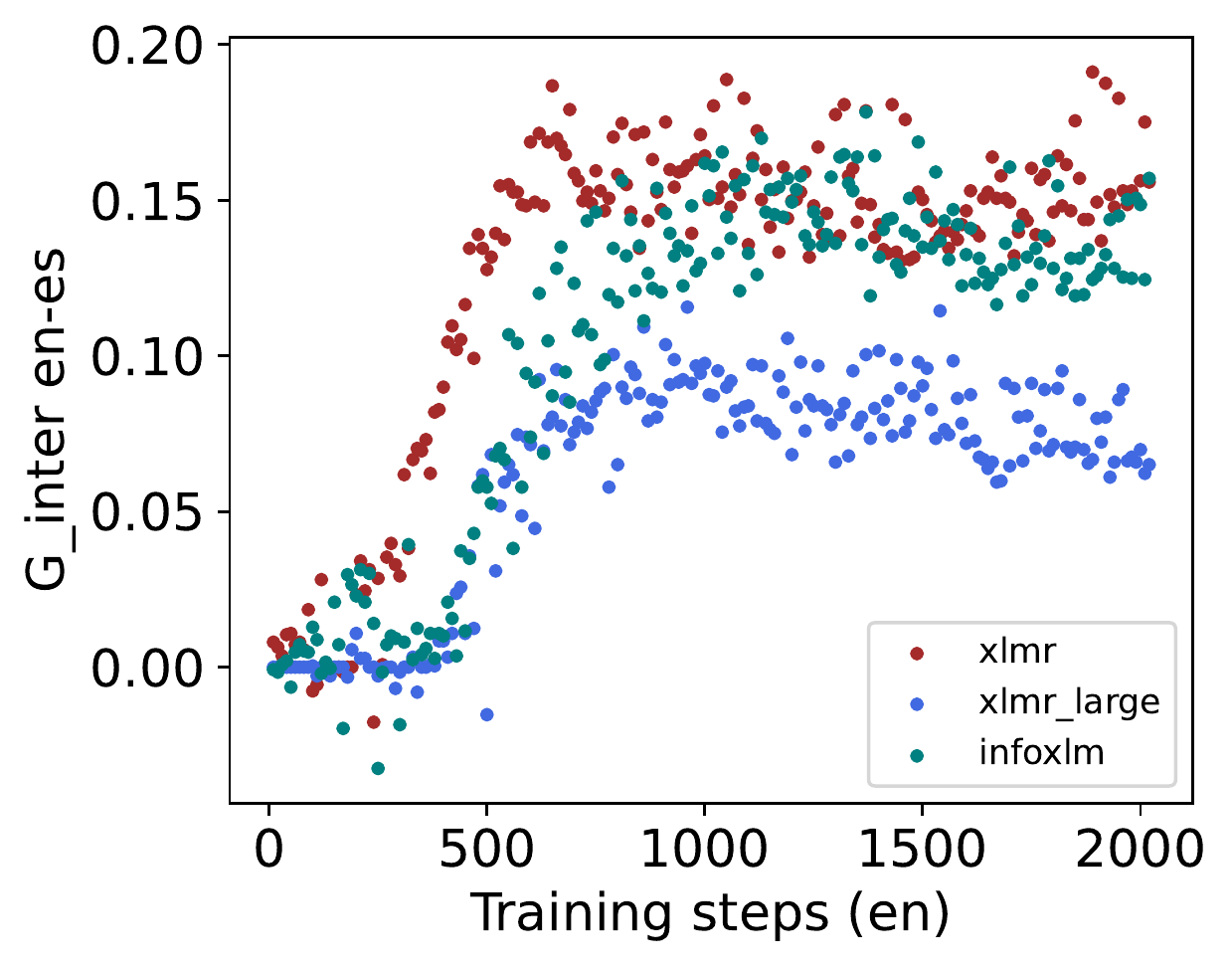}
    \end{subfigure}
    \hfill
    \begin{subfigure}[b]{0.32\textwidth}
     \centering
     \includegraphics[width=0.95\textwidth]{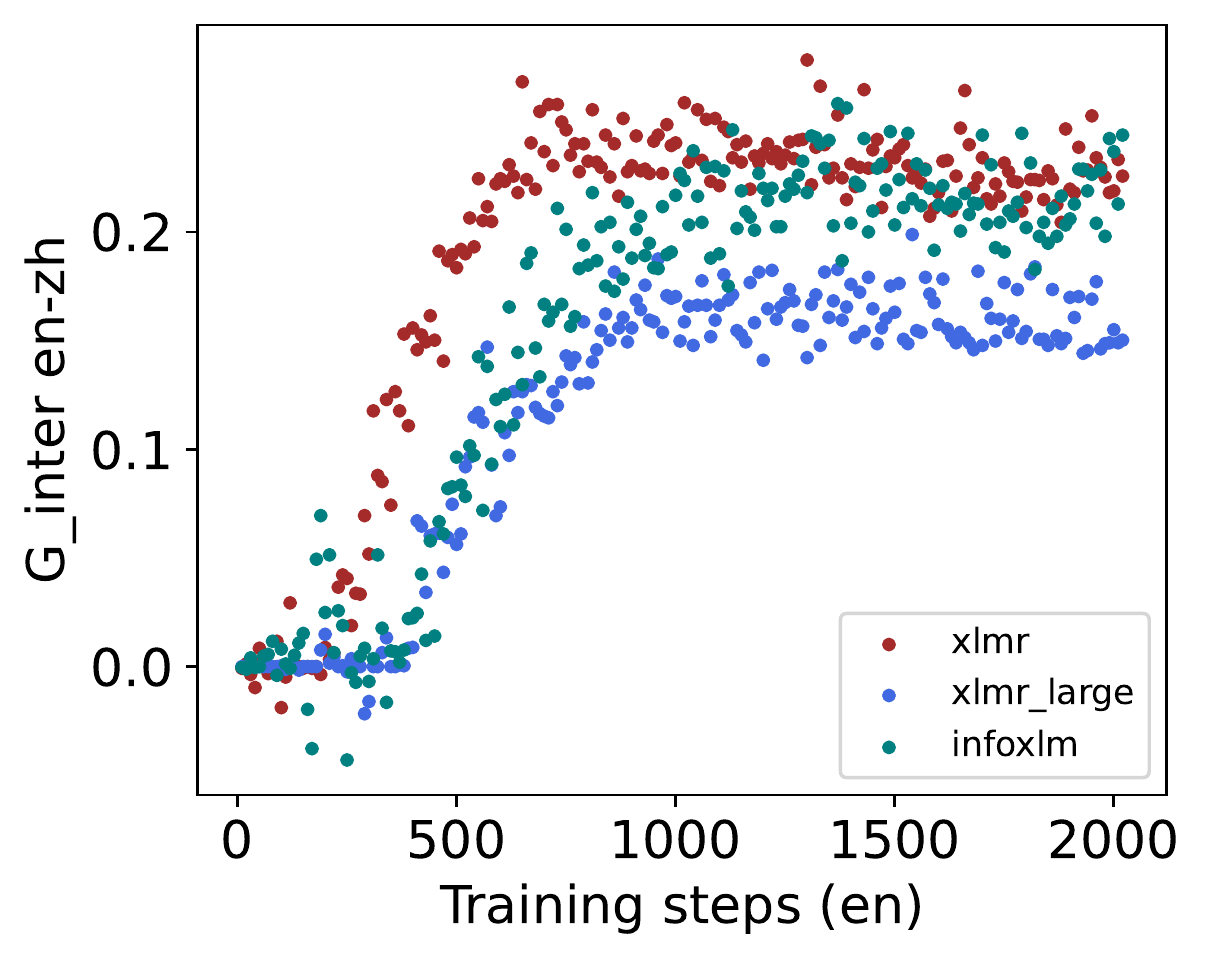}
    \end{subfigure}
    \hfill
    \begin{subfigure}[b]{0.32\textwidth}
     \centering
     \includegraphics[width=0.95\textwidth]{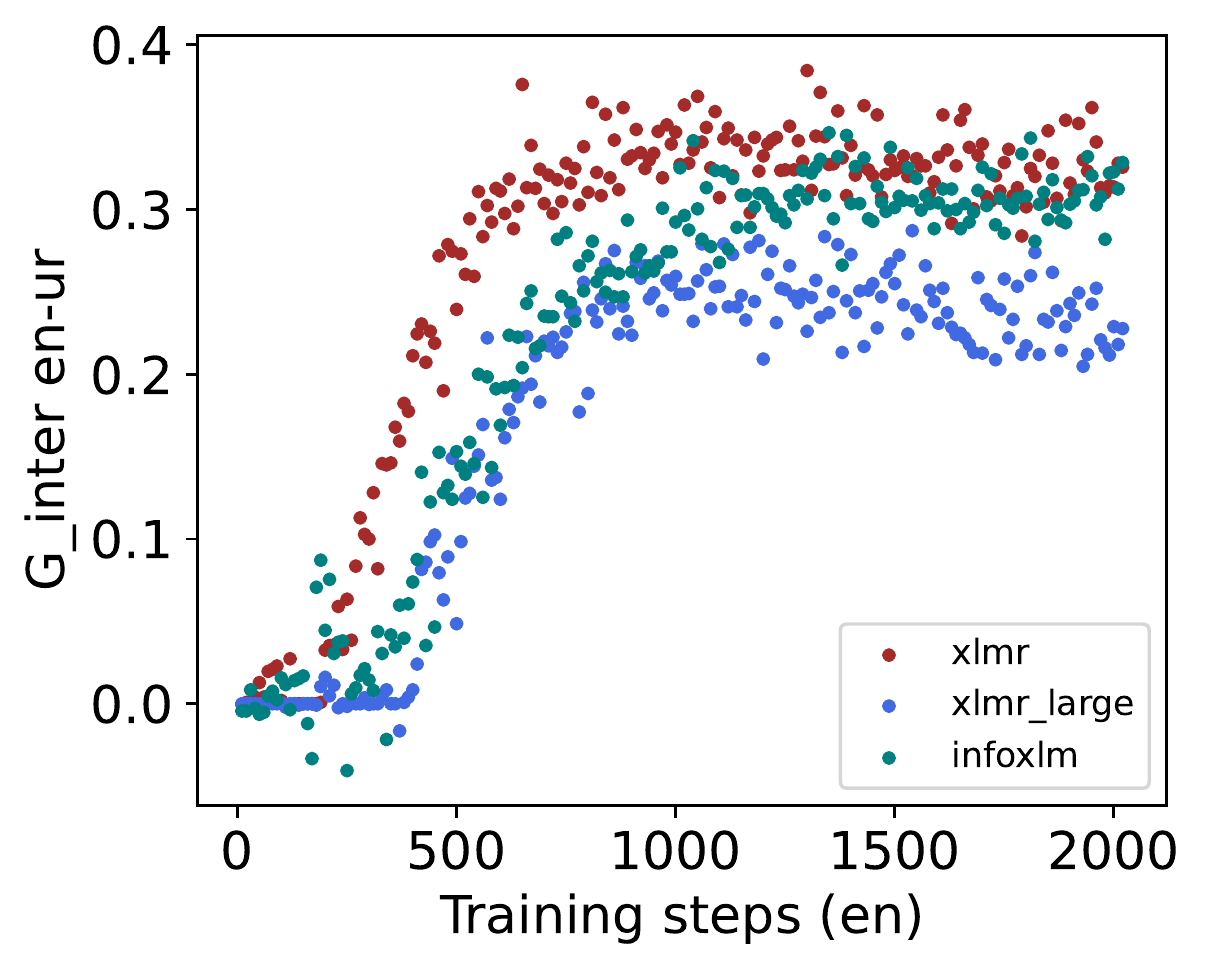}
    \end{subfigure}
\end{subfigure}
\begin{subfigure}[b]{1.0\textwidth}
    \begin{subfigure}[b]{0.32\textwidth}
     \centering
     \includegraphics[width=0.95\textwidth]{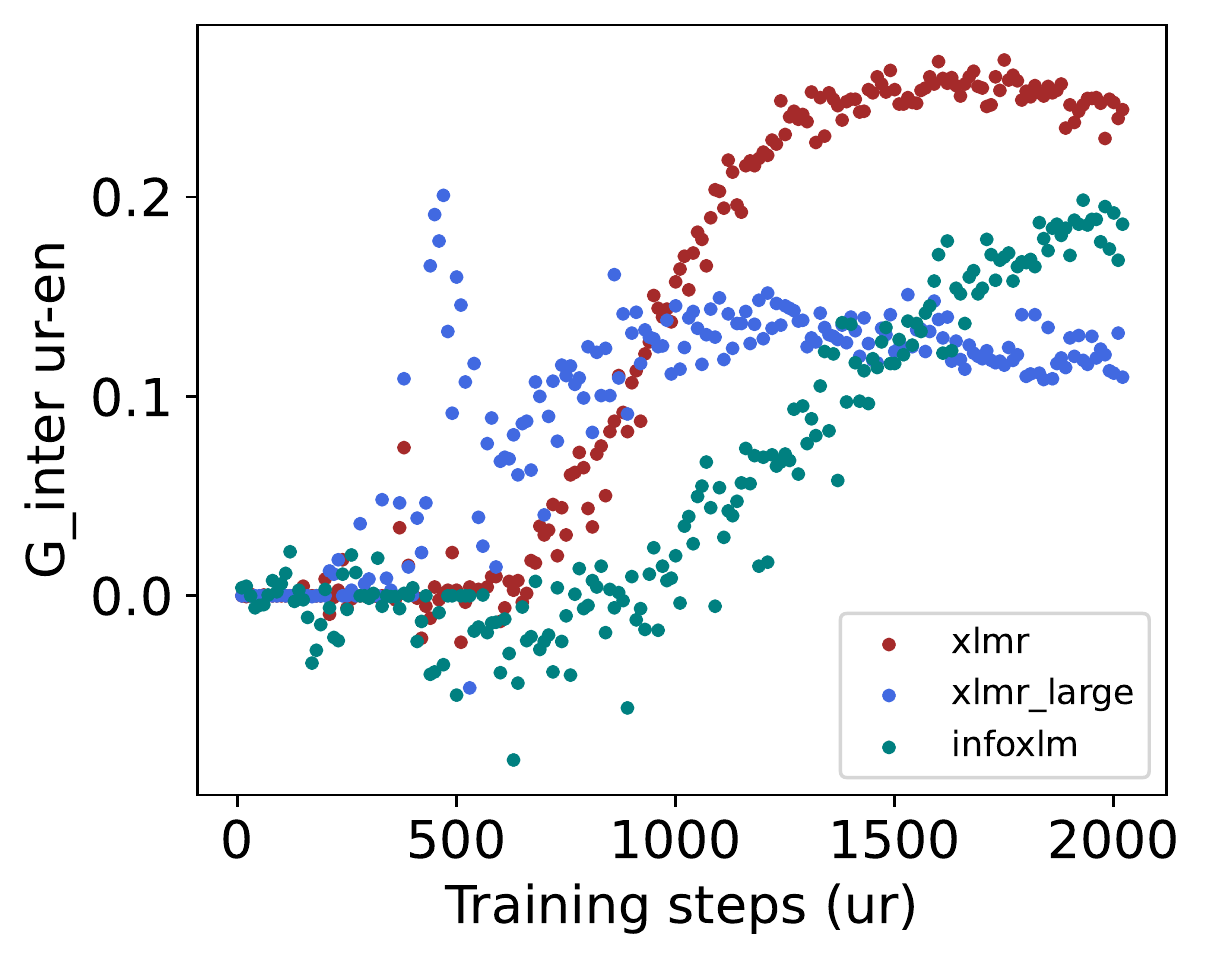}
    \end{subfigure}
    \hfill
    \begin{subfigure}[b]{0.32\textwidth}
     \centering
     \includegraphics[width=0.95\textwidth]{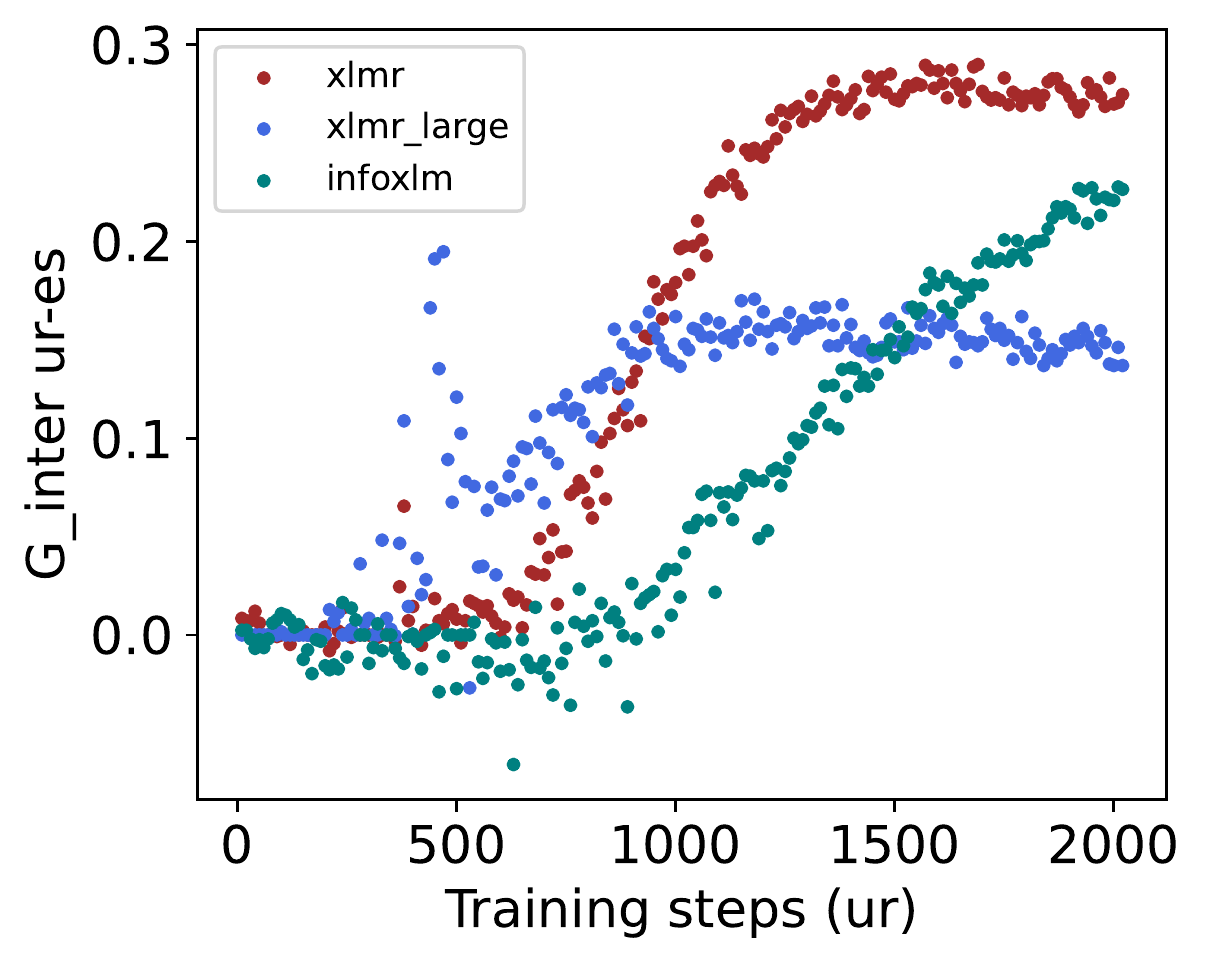}
    \end{subfigure}
    \hfill
    \begin{subfigure}[b]{0.32\textwidth}
     \centering
     \includegraphics[width=0.95\textwidth]{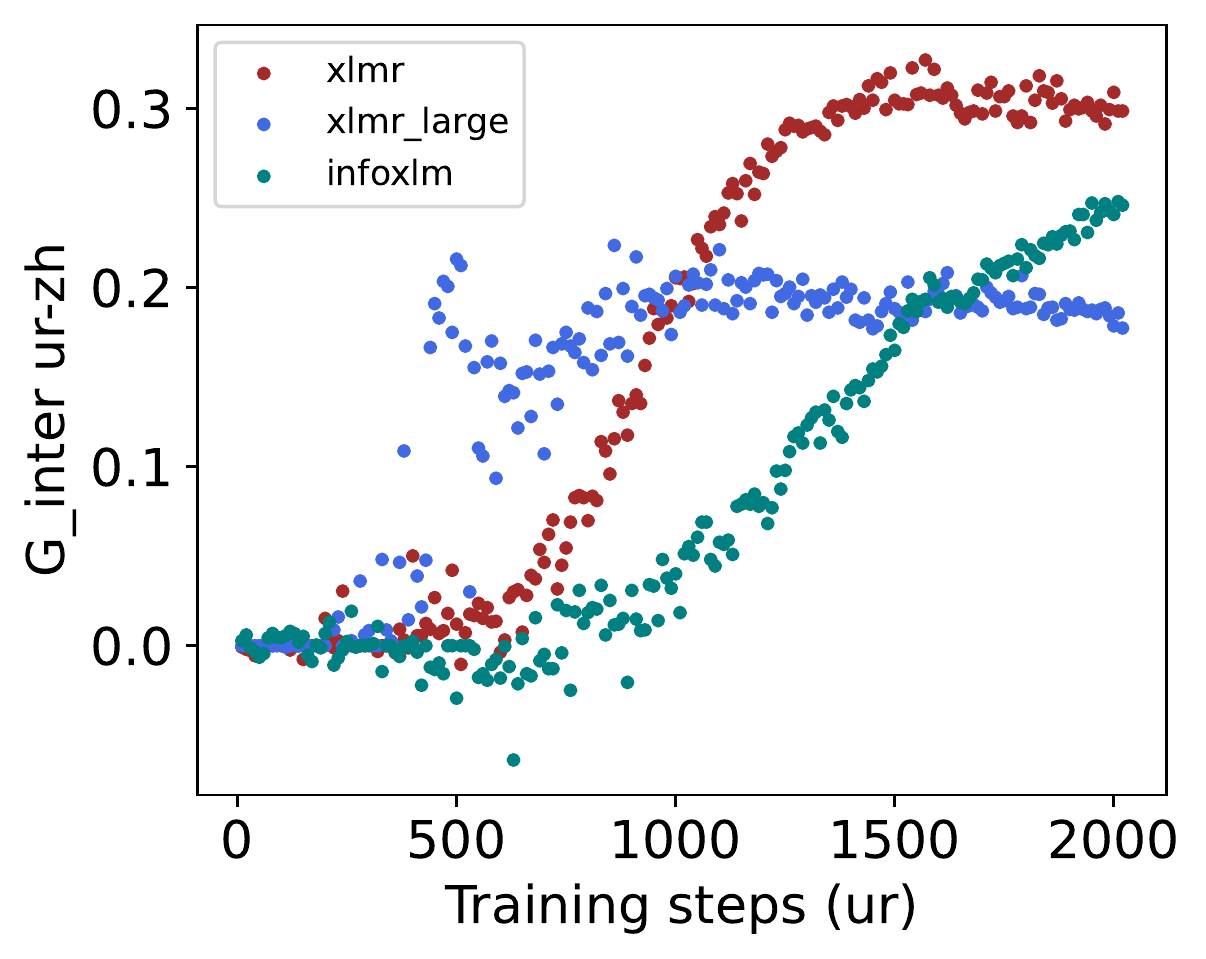}
    \end{subfigure}
    \caption{Interlingual transfer gap scores of various multilingual Transformers.}
\end{subfigure}
\begin{subfigure}[b]{1.0\textwidth}
    \hspace*{\fill}
    \begin{subfigure}[b]{0.32\textwidth}
     \centering
     \includegraphics[width=0.95\textwidth]{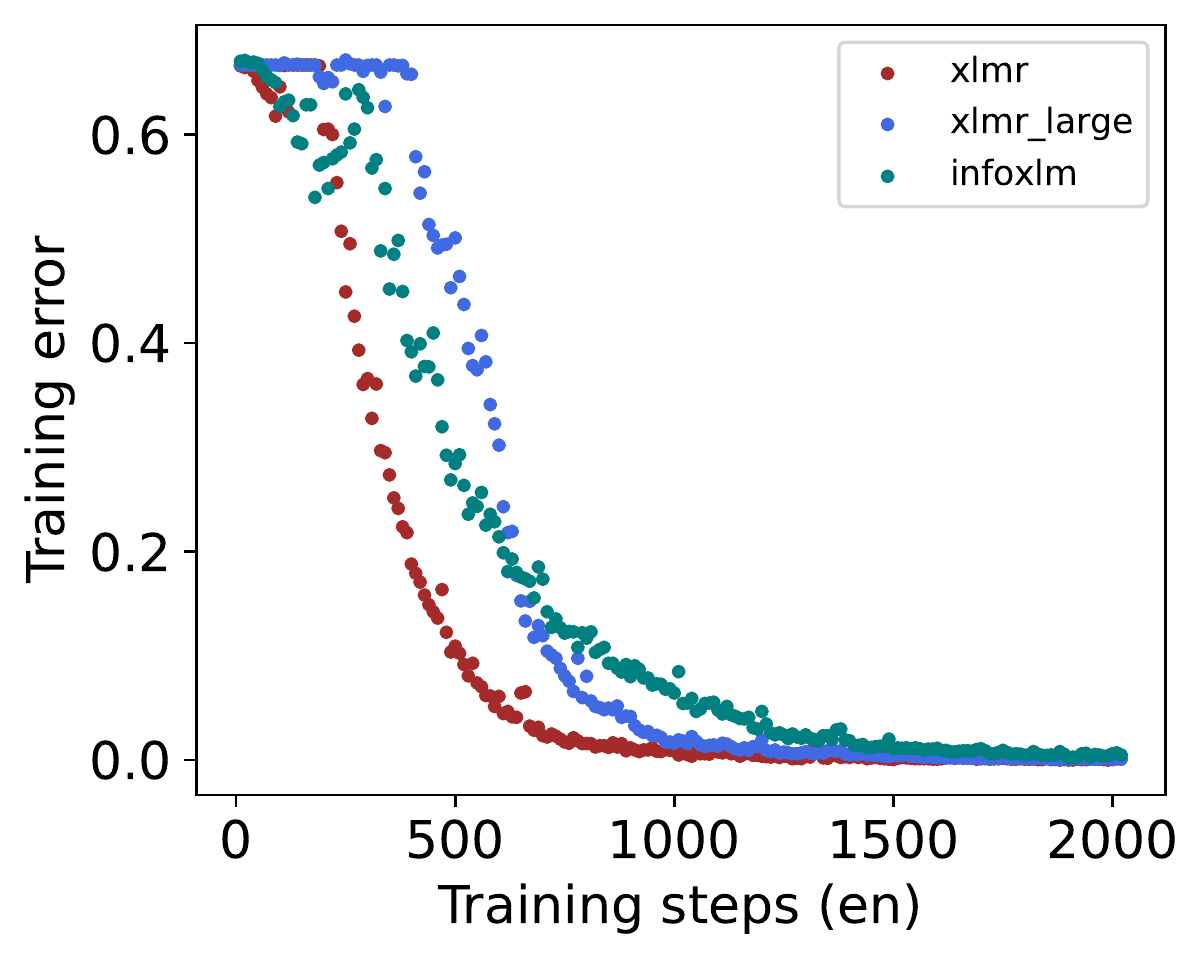}
    \end{subfigure}
    \hfill
    \begin{subfigure}[b]{0.32\textwidth}
     \centering
     \includegraphics[width=0.95\textwidth]{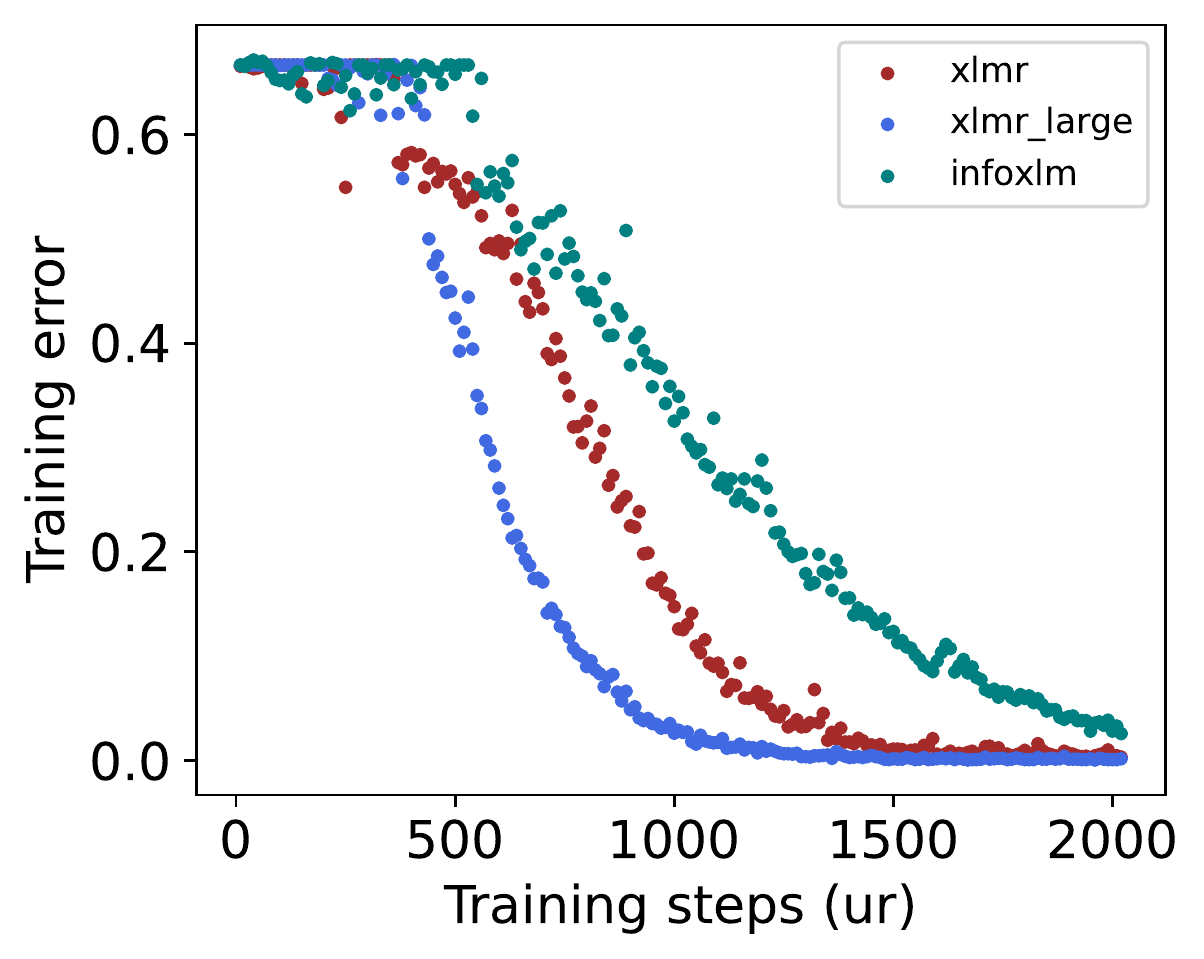}
    \end{subfigure}
    \hspace*{\fill}
    \caption{Training errors of various multilingual Transformers.}
\end{subfigure}
\caption{Intralingual generalization gap, interlingual transfer gap, and training error scores on XNLI natural language inference, where scores from different models are marked in different colors. }
\label{fig:three-comp}
\end{figure*}

\end{document}